\documentclass{article}

\PassOptionsToPackage{numbers, compress}{natbib}
\usepackage[preprint]{neurips_2026}

\usepackage{subcaption}
\usepackage{etoc}
\usepackage[pagebackref=true,colorlinks,linkcolor=blue,citecolor=brown]{hyperref}
\usepackage[utf8]{inputenc} %
\usepackage[T1]{fontenc}    %
\usepackage{hyperref}       %
\usepackage{url}            %
\usepackage{booktabs}       %
\usepackage{amsfonts}       %
\usepackage{amsmath,amssymb}
\usepackage{nicefrac}       %
\usepackage{microtype}      %
\usepackage{xcolor}         %
\usepackage{graphicx}
\usepackage{float}
\usepackage{algorithm}
\usepackage{algorithmic}
\usepackage{multirow}
\usepackage{placeins}
\usepackage{subcaption}
\usepackage{wrapfig}
\usepackage{array}
\usepackage{caption}
\usepackage{enumitem} %
\usepackage{booktabs}
\usepackage{multirow}
\usepackage{graphicx}
\usepackage{capt-of}
\usepackage{adjustbox}
\usepackage{xcolor}

\newcommand{\NEW}[1]{{\color{magenta}[NEW: #1]}}

\title{Temporal Backtracking Search for Test-time Generative Video Reasoning}

\author{
  Sejoon Jun\textsuperscript{1},\
  Zheng Ding\textsuperscript{2},\
  Chloe H. Su\textsuperscript{3},\
  Weirui Ye\textsuperscript{4,$\dagger$},\
  Yilun Du\textsuperscript{3,$\dagger$} \\[3pt]
  \textsuperscript{1}Northeastern University,\
  \textsuperscript{2}Independent Researcher,\
  \textsuperscript{3}Harvard \& Kempner,\
  \textsuperscript{4}MIT \\[3pt]
  $^\dagger$Co-advisors \\
  \url{https://sejoonjun.github.io/TBS/}
}

\newif\ifincludechecklist
\includechecklisttrue

\begin{document}
\maketitle
\etocdepthtag.toc{main}
\begin{abstract}
While test-time scaling has revolutionized reasoning in large language models, generative video reasoning remains bottlenecked by a single-shot paradigm. We demonstrate that searching over denoising steps cannot rescue logically flawed rollouts because spatial trajectories commit early in the diffusion process. Root-level Best-of-$N$ (BoN) sampling is similarly inefficient: reasoning errors cluster early in the temporal axis, and resampling blindly discards verified upstream progress. To unlock effective test-time scaling for video models, we introduce \textbf{Temporal Backtracking Search (TBS)}, which shifts the search space to the temporal axis. TBS transforms video generation into an iterative generate--verify--restart loop via three core mechanisms: (1) \textit{variable-$K$ conditioning} to resume generation from arbitrary clean prefixes; (2) \textit{temporal process verification} to localize failures and extract valid restart anchors; and (3) \textit{prefix-based search} to reallocate compute toward extending correct trajectories rather than root resampling. Across algorithmic, navigation, and robotics domains, TBS Pareto-dominates matched-budget BoN. In a strict out-of-distribution setting where one-shot generation collapses ($0.7\%$ for BoN), TBS achieves $22.7\%$, with every solved episode stemming from a restarted branch. Ultimately, TBS reveals that the local reasoning competence of video models far exceeds what single-shot rollouts indicate, providing a scalable test-time framework to unlock it.
\end{abstract}

\section{Introduction}
\label{sec:intro}

Generative video models are evolving beyond visual content creation to emerge as world simulators for sequential decision-making. By extracting action plans from generated rollouts, agents can search, plan, and solve complex tasks entirely in imagination~\citep{du2023learning, bruce2024genie, brooks2024video, yang2023foundation, ha2018world, lecun2022path}. In this reasoning-oriented regime, the relevant axis for scaling test-time compute is no longer aesthetic quality but \textit{trajectory correctness}---whether the generated video depicts a logically valid solution to the given problem. 

\begin{figure}[t]
\vskip -.5cm
\centering
    \includegraphics[width=\textwidth]{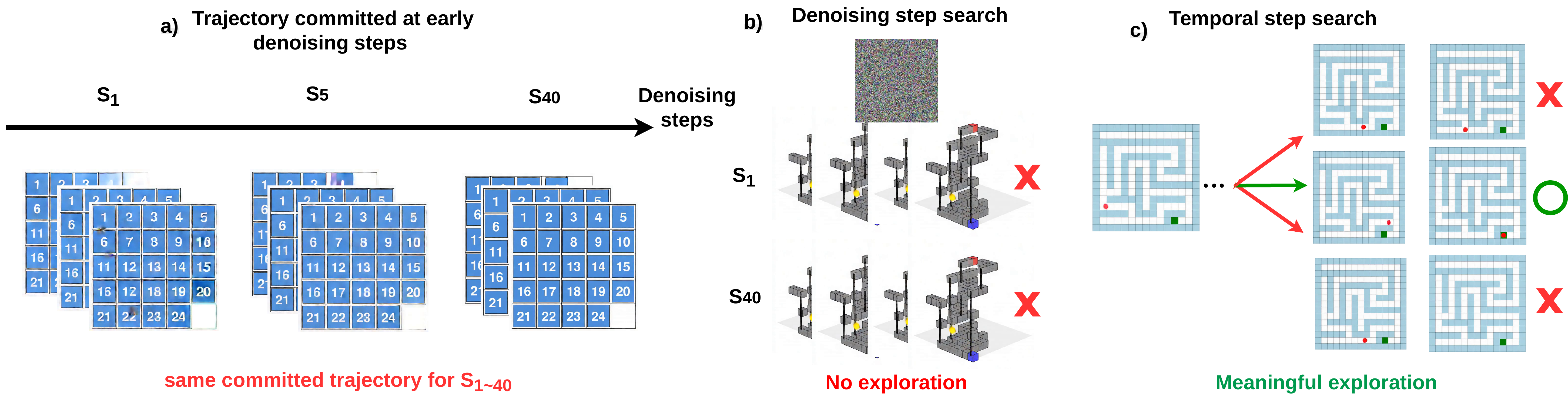}
\caption{\textbf{Trajectories are determined by the earliest denoising steps}. Decoding the same noisy latent at successive denoising steps shows that the motion plan is fixed within the first few steps and persists through to the final rollout. \textbf{Takeaway}: Additional compute spent on later denoising steps cannot redirect a wrong trajectory, so test-time search for video reasoning must branch over temporal frames, not over denoising steps.}
\label{fig:main}
\vskip -.2cm
\end{figure}

While recent test-time scaling methods for video generation optimize over noise schedules or partially denoised latents to improve visual fidelity and prompt alignment~\citep{dlbs, evosearch, yang2025scalingnoise, cong2025can, liu2025video, wu2026imagerysearch}, they fall short for logical decision-making. Consequently, for temporal scaling, the community typically defaults to Best-of-$N$ (BoN) sampling, treating an entire video rollout as an indivisible temporal block.

We identify two empirical bottlenecks that cause these paradigms to fail in long-horizon reasoning. First, decoding intermediate latents reveals that video models commit to a high-level spatial trajectory within the earliest denoising steps (Figure~\ref{fig:main}); allocating test-time compute to later denoising steps therefore cannot rescue a trajectory that is logically flawed from the start. Second, reasoning failures are not uniformly distributed along the trajectory; they cluster heavily in the early stages of generation. Root-level BoN thus wastes compute by repeatedly resampling the same early failure regions, discarding the valid partial prefixes generated upstream.

These observations motivate shifting the search space from the denoising axis to the \textit{temporal} axis. Instead of discarding an entire rollout upon a single mistake, search algorithms should isolate the failure, salvage verified preceding frames, and reallocate compute to repair the invalid suffix. We introduce \textbf{Temporal Backtracking Search (TBS)}, transforming single-shot video generation into an iterative generate--verify--restart loop. TBS realizes prefix-based temporal search via three core components: (1) \textit{variable-$K$ conditioning}, a curriculum-trained capability that enables the diffusion model to resume from arbitrary clean prefixes while preserving prior motion context~\citep{chen2024diffusion, sun2025ar, xie2025progressive}; (2) \textit{temporal process verification}, which identifies the first failure frame to extract verified restart anchors~\citep{lightman2023let, uesato2022solving, setlur2024rewarding}; and (3) \textit{prefix-based search}, which reallocates inference compute to extend valid trajectories rather than resampling from the root~\citep{zhang2025inference, ma2025inference}.

We evaluate TBS across three regimes spanning the realistic spectrum of process supervision: exact symbolic verification in algorithmic grid-worlds, learned noisy trajectory prediction in continuous navigation~\citep{wang2025target}, and simulator-replay verification in robotic manipulation~\citep{chen2025robotwin, mu2024robotwin}. Across all three domains, TBS Pareto-dominates matched-budget BoN. Notably, in a strict out-of-distribution (OOD) setting where one-shot generation collapses ($0.7\%$ exact match for BoN), TBS achieves $22.7\%$, with every solved episode stemming from a restarted child branch. Our primary contributions are:
\begin{itemize}
\item Identifying two critical bottlenecks in video test-time scaling: early trajectory commitment renders denoising-step search ineffective, and clustered early failures make root-level BoN grossly inefficient.
\item Proposing \textbf{Temporal Backtracking Search (TBS)}, which transitions video test-time scaling from monolithic single-shot sampling to prefix-based temporal correction.
\item Demonstrating empirically that temporal backtracking unlocks and stitches together the underlying local reasoning competence of video models under symbolic, learned, and physics-based verification.
\end{itemize}

\section{Related Work}
\label{sec:related}

\textbf{Test-time scaling and process verification.} Test-time scaling via Best-of-$N$ and tree search has transformed LLM reasoning~\citep{snell2024scaling, brown2024large, feng2024alphazero}, driven by the insight that \emph{process} verifiers scoring intermediate steps strictly outperform \emph{outcome} verifiers scoring only final answers~\citep{lightman2023let, uesato2022solving, setlur2024rewarding}. TBS adapts this paradigm to generative video reasoning by shifting evaluation from monolithic rollouts to verified prefixes. Crucially, our process verifier localizes \emph{where} a trajectory deviates, rather than merely judging \emph{whether} the terminal sequence is flawed.

\textbf{Long-horizon video generation.} Diffusion Forcing~\citep{chen2024diffusion} demonstrates causal continuation from clean prefixes as a native sequence diffusion capability, providing the modeling primitive for our variable-$K$ conditioning. While generative stitching methods~\citep{luo2025generative, song2025generative} also compose long horizons from local segments, they heavily rely on bidirectional harmonization or global joint resampling. In contrast, TBS performs causal, verifier-localized repair, leaving verified pasts strictly untouched.

\textbf{Test-time scaling for video generation.} Recent methods scale inference compute for video diffusion by searching over denoising trajectories or partially denoised latents~\citep{liu2025video, dlbs, evosearch, yang2025scalingnoise, cong2025can, wu2026imagerysearch}. However, these approaches operate entirely on the \emph{denoising} axis to optimize visual aesthetics and prompt alignment. TBS fundamentally differs by searching the \emph{temporal} axis to optimize trajectory correctness, leveraging process supervision that ranges from symbolic rules to simulator replay. Furthermore, while concurrent studies~\citep{wang2026demystifying, wiedemer2025video} diagnose the long-horizon reasoning limits of video models, they offer no constructive mitigation---a critical gap TBS bridges via prefix-based restartable generation.

\section{Method}
\label{sec:method}

\begin{figure}[t]
\centering
\vskip -.5cm
\includegraphics[width=\textwidth]{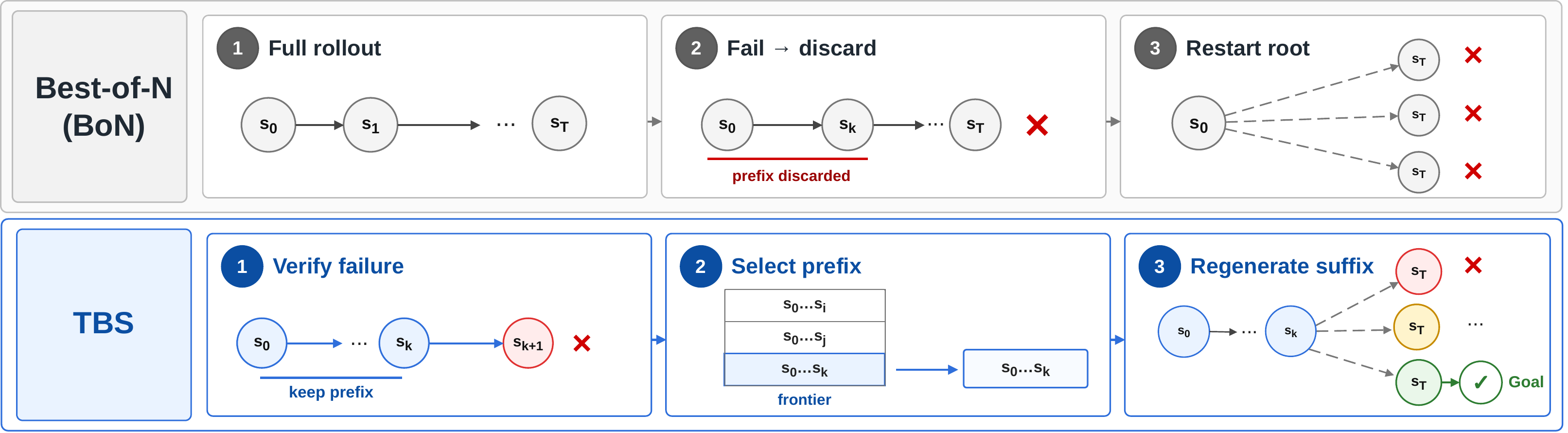}
\caption{\textbf{Mechanism of Temporal Backtracking Search (TBS).}
(a)~\textbf{Single-shot generation:} The trajectory is sampled in one pass; a single error at frame $F$ invalidates the entire remaining suffix.
(b)~\textbf{TBS:} Decomposes long-horizon reasoning into short-horizon sub-problems by iteratively extending verified prefixes. The algorithm executes a generate--verify--restart loop: 1)~\textbf{Generate} a candidate rollout; 2)~\textbf{Verify} the trajectory to localize the first failure and cache the valid prefix; and 3)~\textbf{Restart} generation from the highest-priority anchor.}
\label{fig:overview}
\vskip -.3cm
\end{figure}

To break the single-shot generative bottleneck, we reformulate video reasoning as inference-time search over the temporal axis. This section formalizes the \textit{conditional reuse gap} that drives our approach (Section~\ref{sec:problem}), introduces the \textbf{Temporal Backtracking Search (TBS)} algorithm designed to exploit this gap (Section~\ref{sec:tbs}), and details the variable-$K$ generative engine required to enable temporal branching (Section~\ref{sec:variable_k}).

\subsection{Problem Formulation and the Conditional Reuse Gap}
\label{sec:problem}

Let $\mathbf{v} = (v_1, \ldots, v_T)$ be a video of $T$ frames generated from an initial image $v_1$ and a text prompt $p$. For sequential reasoning tasks, we define the logical correctness of a trajectory via a transition validity function:
\begin{equation}
\phi(v_1, \ldots, v_t) = \mathbb{I}\!\left[\,v_i \to v_{i+1} \text{ is valid } \forall\, i \le t{-}1\,\right].
\label{eq:validity}
\end{equation}
A video achieves an \emph{exact match} (EM\,$=1$) if and only if $\phi(\mathbf{v}) = 1$ and the trajectory reaches the goal state by frame $T$. For invalid rollouts, we identify the earliest logical failure as the \emph{first failure frame}:
\begin{equation}
F(\mathbf{v}) = \min\{t : \phi(v_1, \ldots, v_t) = 0\},
\label{eq:failure}
\end{equation}
with $F(\mathbf{v}) = T{+}1$ if the video is fully correct. This failure index naturally partitions the rollout into a valid prefix $\mathbf{v}_{1:F-1}$ and an invalid suffix.

Root-level scaling paradigms, such as BoN, indiscriminately discard this verified prefix upon encountering a failure. To illustrate why this is suboptimal, consider a ground-truth trajectory $\mathbf{v}^*$ where a generated sample matches $\mathbf{v}^*$ up to frame $\ell$. A root-level generator must synthesize the entire future trajectory from the initial frame, $P_{\mathrm{root}} = p_\theta(\mathbf{v}^*_{2:T} \mid v_1^*, p)$, whereas prefix repair only needs to resolve the remaining suffix, $P_{\mathrm{repair}}(\ell) = p_\theta(\mathbf{v}^*_{\ell+1:T} \mid \mathbf{v}^*_{1:\ell}, p)$.

We define the discrepancy in modeling difficulty between $P_{\mathrm{root}}$ and $P_{\mathrm{repair}}$ as the \textbf{conditional reuse gap}. Because reasoning errors accumulate multiplicatively over time, extending a clean partial sequence is inherently more tractable than generating a long-horizon rollout from scratch. TBS explicitly exploits this gap.

\begin{algorithm}[t]
\small
\caption{Temporal Backtracking Search (TBS)}
\label{alg:tbs}
\begin{algorithmic}[1]
\REQUIRE Initial frame $v_1$, prompt $p$, budgets $(k_1, s, m)$, Verifier $\mathcal{V}$, Generator $\mathcal{G}$
\STATE Initialize priority frontier $\mathcal{F} \leftarrow \emptyset$ \COMMENT{Pool of verified prefix anchors $(K, \mathbf{v}_{1:R}, \pi)$}
\STATE Sample $k_1$ root rollouts from $\mathcal{G}$ conditioned on $(K{=}1, [v_1], p)$
\FOR{each generated video $\mathbf{v}$}
  \STATE $(\text{EM}, F, R, K, \pi) \leftarrow \mathcal{V}(\mathbf{v})$ \COMMENT{$F$: first failure frame; $R$: aligned restart anchor; $\pi$: verifier-assigned priority}
  \IF{$\text{EM}{=}1$}
    \RETURN $\mathbf{v}$
  \ELSIF{$R > 1$}
    \STATE $\mathcal{F} \leftarrow \mathcal{F} \cup \{(K, \mathbf{v}_{1:R}, \pi)\}$ \COMMENT{Store verified prefix as a new anchor}
  \ENDIF
\ENDFOR
\FOR{expansion step $j = 2, \ldots, m$}
  \STATE $n^* \leftarrow \arg\max_{n \in \mathcal{F}} \pi_n$ \COMMENT{Select highest-priority anchor}
  \STATE $\mathcal{F} \leftarrow \mathcal{F} \setminus \{n^*\}$
  \STATE Sample $s$ child suffixes from $\mathcal{G}$ conditioned on prefix $n^*$
  \STATE Evaluate children with $\mathcal{V}$; insert unsolved valid prefixes into $\mathcal{F}$
  \IF{any child achieves $\text{EM}{=}1$}
    \RETURN solved child video
  \ENDIF
\ENDFOR
\RETURN best unsolved video by priority $\pi$
\end{algorithmic}
\end{algorithm}

\subsection{Temporal Backtracking Search (TBS)}
\label{sec:tbs}

Leveraging the conditional reuse gap, TBS performs state-space search directly over temporal prefixes. The search maintains a \textbf{priority frontier} $\mathcal{F}$, a pool of verified prefix anchors awaiting expansion. Each entry in $\mathcal{F}$ is a tuple $(K, \mathbf{v}_{1:R}, \pi)$, where $\mathbf{v}_{1:R}$ is the verified prefix, $K = R/\tau$ is its corresponding latent length (Section~\ref{sec:variable_k}), and $\pi \in \mathbb{R}$ is a deterministic priority score assigned by the verifier (higher $\pi$ indicates a more promising anchor for continuation). Instead of repeatedly querying the computationally expensive $P_{\mathrm{root}}$, TBS isolates localized failures and samples continuations strictly from anchors stored in $\mathcal{F}$.

Guided by a process-level verifier $\mathcal{V}$ (Algorithm~\ref{alg:tbs}), the search proceeds as follows. For each rollout, $\mathcal{V}$ evaluates the exact-match status and extracts the first failure frame $F$, an aligned restart anchor $R \le F{-}1$, and the priority score $\pi$. If an invalid rollout contains reusable progress ($R > 1$), the corresponding tuple is inserted into $\mathcal{F}$. At each expansion step, TBS pops the highest-priority anchor from $\mathcal{F}$ and queries the generator for $s$ candidate suffix continuations.

For latent diffusion models, the restart frame must strictly align with the VAE's temporal latent factor $\tau$. The aligned restart anchor is computed as $R(\mathbf{v}) = \tau \cdot \lfloor (F(\mathbf{v}) - 1) / \tau \rfloor$. This guarantees the generator cleanly resumes from exactly $K = R/\tau$ latent frames, preventing temporal overlap or information leakage between the clean prefix and the resampled suffix.

\subsection{Variable-$K$ Conditioning}
\label{sec:variable_k}

For TBS to function, the generator must support temporal branching: given a prefix verified up to frame $R$, the model must synthesize the suffix while preserving the exact prefix conditioning. Standard video diffusion models lack this capability, as they denoise all frames jointly from a single-frame initial condition ($K{=}1$).

To unlock temporal branching while keeping per-branch inference cost constant, we finetune the video diffusion model (based on Wan2.2-TI2V-5B\cite{wan2025}) with          
\textbf{variable-$K$ conditioning}. Let $\mathbf{z} = \mathcal{E}(\mathbf{v}) \in \mathbb{R}^{C \times T_z \times H_z
\times W_z}$ be the VAE-encoded latent representation. During training, we apply a per-frame timestep mask: the first $K$ 
latent frames remain clean ($t_i = 0$), while the remaining suffix frames are noised at a single sampled timestep $t$:
\begin{equation}
\tilde{\mathbf{z}}_i =                                                                                        
\begin{cases}
\mathbf{z}_i & i \leq K, \\                                                                                      
(1 - t)\,\mathbf{z}_i + t\,\boldsymbol{\epsilon}_i,\quad
\boldsymbol{\epsilon}_i \sim \mathcal{N}(0,\mathbf{I}) & i > K,                                                           
\end{cases}                                                                                                               
\label{eq:variable_k_mask}                                                                                                
\end{equation}                                                                                                            
At inference, $K$ is dictated by the verifier's restart anchor ($K =      
R/\tau$). We finetune the base video diffusion model with a progressive curriculum that shifts the training distribution
from $K{=}1$ to a broad mixture up to $K{=}20$. Without this curriculum, the model collapses when conditioned on longer
prefixes (Appendix~\ref{app:training}).

A naive branching alternative is extracting the last verified frame to use as a $K{=}1$ initial condition. However, a single static frame strips away the motion, dynamics, and task momentum established by the preceding trajectory. To isolate the necessity of motion context, we evaluated a controlled subset of 40 provably repairable long-horizon Maze trajectories. Replacing the full temporal prefix with a single-frame state reset collapsed the repair success rate from $100\%$ to $2.5\%$ (Appendix~\ref{app:prefix_repair}). Full prefix conditioning is therefore the foundational engine enabling temporal backtracking.

\section{Experiments}
\label{sec:experiments}
\providecommand{\NEW}[1]{{\color{magenta}[NEW: #1]}}

\begin{figure}[t]
\vskip -.5cm
  \centering
  \begin{subfigure}[b]{\textwidth}
    \centering
    \includegraphics[width=\linewidth, height=0.8\textheight, keepaspectratio]{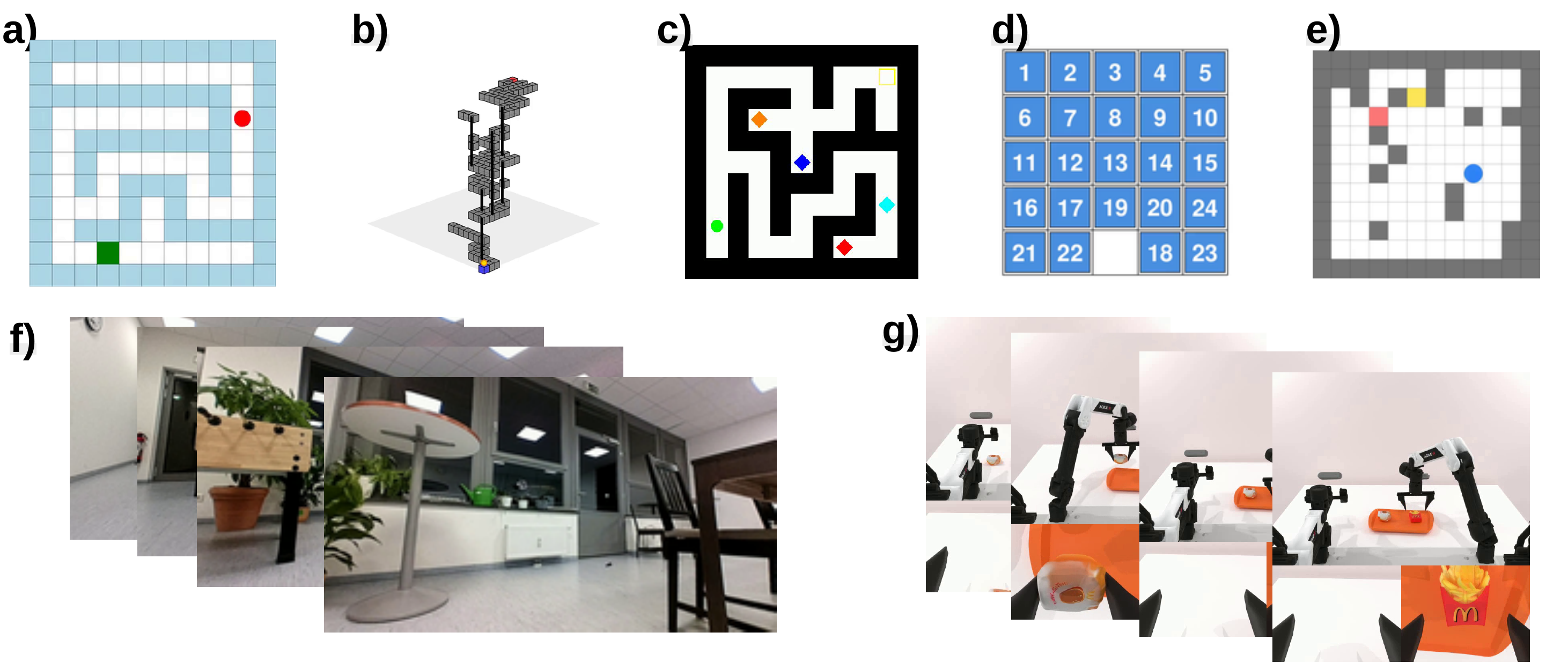}
  \end{subfigure}
  \caption{\textbf{Evaluation domains.} Rule-based grid-worlds: \textbf{2D Maze} (a), \textbf{3D Maze} (b), \textbf{Keys-and-Doors} (c), \textbf{Sliding Puzzle} (d), \textbf{Sokoban} (e); continuous navigation \textbf{Target-Bench} (f); robotic manipulation \textbf{RoboTwin} (g).}
  \label{fig:domains}
  \vskip -.1cm
\end{figure}

\subsection{Experimental Setup}
\label{sec:setup}

We evaluate TBS in three regimes spanning the spectrum of process supervision available in practice: exact symbolic verification in rule-based grid-worlds, learned trajectory prediction in continuous navigation, and simulator-replay verification in robotic manipulation.

\noindent\textbf{Structured video reasoning.}
We use five rule-based grid-worlds (Figure~\ref{fig:domains}), where deterministic game rules serve as verifiers and identify the first failure frame. The domains target distinct reasoning bottlenecks: \textit{2D Maze} and \textit{3D Maze} for spatial path adherence, \textit{Sliding Puzzle} for semantic correctness, \textit{Sokoban} for irreversible dead states, and \textit{Keys-and-Doors} for long-horizon sub-goal ordering. Each domain has $600$ evaluation samples split into easy/medium/hard; \textit{Keys-and-Doors} additionally includes a $300$-sample out-of-distribution (OOD) split with extended trajectories (Appendix~\ref{app:domains},~\ref{app:verifier}).

\noindent\textbf{Continuous navigation.}
We use Target-Bench~\citep{wang2025target}, a mapless path-planning benchmark whose official score (Eq.~\ref{eq:targetbench_score}) aggregates five trajectory metrics against a ground-truth path:
\begin{equation}
S = 0.05\,e^{-\mathrm{ADE}}
+ 0.10\,e^{-\mathrm{FDE}}
+ 0.10\,(1{-}\mathrm{MR}/100)
+ 0.65\,\mathrm{SE}\cdot\mathrm{AC},
\label{eq:targetbench_score}
\end{equation}
where ADE and FDE are the average and final $L_2$ displacement errors against the GT, MR is the percentage of trajectory points farther than $1$\,m from the GT, SE and AC are the benchmark's soft-endpoint and approach-consistency scores, and TC is the trajectory-coverage term (full definitions in Appendix~\ref{app:targetbench}). The verifier is a learned trajectory predictor that maps the current frame and prompt to a future metric-space path, and TBS branches when the decoded video trajectory leaves the predicted spatial corridor.

\noindent\textbf{Robotic manipulation.}
We evaluate manipulation on RoboTwin~\citep{mu2024robotwin}, a bimanual robot benchmark backed by a physics simulator. Generated videos are decoded into 14-D action sequences via an inverse dynamics model and rolled out in the simulator, whose ground-truth task predicate serves as the verifier. RoboTwin trajectories have a natural two-phase structure---an approach phase followed by a final goal-attempt window---which TBS uses to choose restart anchors (Section~\ref{sec:mechanism}).

\noindent\textbf{Baseline and compute budget.}
Our baseline is Best-of-$N$ (BoN), the standard test-time-scaling method for video generation~\citep{wiedemer2025video, wang2026demystifying}. To isolate the effect of prefix reuse, both methods share identical terminal evaluators and root candidate pools. A TBS configuration uses root width $k_1$, child samples $s$, and maximum expansions $m$, giving budget $B = k_1 + s\,(m{-}1)$; we compare against BoN at matched budget $N{=}B$. The \textit{Time} columns in Table~\ref{tab:main_structured_results_pm} report normalized wall-clock cost, which grows sub-linearly due to early termination after the first verified solve.

\subsection{Main Results}
\label{sec:main_results}

\begin{table*}[t]
\vskip -.5cm
\caption{\textbf{Structured video reasoning, exact-match results.} Reported as exact-match percentage with $95\%$ Wilson confidence intervals; intervals are clipped at $0$ and $100$. Bold marks the better method; ties are not bolded. The TBS--BoN gap widens on \textit{Hard} and \textit{Strict-OOD} splits.}
\label{tab:main_structured_results_pm}
\centering
\footnotesize
\setlength{\tabcolsep}{5pt}
\renewcommand{\arraystretch}{0.95}
\begin{tabular}{ll cc cc cc}
\toprule
\multirow{2}{*}{\textbf{Domain}} & \multirow{2}{*}{\textbf{Split / Metric}} & \multicolumn{2}{c}{\textbf{Baselines}} & \multicolumn{2}{c}{\textbf{Budget 10}} & \multicolumn{2}{c}{\textbf{Budget 20}} \\
\cmidrule(lr){3-4} \cmidrule(lr){5-6} \cmidrule(lr){7-8}
& & \textbf{BoN-1} & \textbf{BoN-5} & \textbf{BoN-10} & \textbf{TBS-10} & \textbf{BoN-20} & \textbf{TBS-20} \\
\midrule
\multirow{5}{*}{Maze}
& Time           & $1.00{\times}$ & $2.58{\times}$ & $4.01{\times}$ & $3.53{\times}$ & $6.25{\times}$ & $4.42{\times}$ \\
& Overall        & 51.0$\pm$4.0 & 67.7$\pm$3.9 & 75.2$\pm$3.6 & \textbf{84.0$\pm$3.1} & 79.0$\pm$3.4 & \textbf{90.3$\pm$2.6} \\
& Easy           & 98.0$\pm$2.0 & 100.0$_{-1.9}$ & 100.0$_{-1.9}$ & 100.0$_{-1.9}$ & 100.0$_{-1.9}$ & 100.0$_{-1.9}$ \\
& Med.           & 76.5$\pm$6.3 & 93.0$\pm$4.4 & 96.0$\pm$3.7 & \textbf{99.0$\pm$2.6} & 97.0$\pm$3.4 & \textbf{99.5$\pm$2.3} \\
& Hard           & 18.3$\pm$4.8 & 40.0$\pm$5.6 & 53.0$\pm$5.7 & \textbf{68.7$\pm$5.5} & 60.0$\pm$5.6 & \textbf{81.0$\pm$4.8} \\
\midrule
\multirow{5}{*}{Maze3D}
& Time           & $1.00{\times}$ & $1.38{\times}$ & $1.59{\times}$ & $1.57{\times}$ & $1.95{\times}$ & $1.83{\times}$ \\
& Overall        & 85.5$\pm$3.0 & 95.3$\pm$2.0 & 96.2$\pm$1.9 & \textbf{97.2$\pm$1.7} & 96.7$\pm$1.8 & \textbf{97.7$\pm$1.6} \\
& Easy           & 100.0$_{-1.9}$ & 100.0$_{-1.9}$ & 100.0$_{-1.9}$ & 100.0$_{-1.9}$ & 100.0$_{-1.9}$ & 100.0$_{-1.9}$ \\
& Med.           & 97.0$\pm$3.4 & 99.5$\pm$2.3 & 99.5$\pm$2.3 & \textbf{100.0$_{-1.9}$} & 99.5$\pm$2.3 & \textbf{100.0$_{-1.9}$} \\
& Hard           & 73.0$\pm$5.3 & 91.0$\pm$3.8 & 92.7$\pm$3.6 & \textbf{94.3$\pm$3.2} & 93.7$\pm$3.4 & \textbf{95.3$\pm$3.0} \\
\midrule
\multirow{5}{*}{Sokoban}
& Time           & $1.00{\times}$ & $2.78{\times}$ & $4.17{\times}$ & $4.19{\times}$ & $6.29{\times}$ & $6.20{\times}$ \\
& Overall        & 41.5$\pm$4.0 & 67.5$\pm$3.8 & 75.5$\pm$3.6 & \textbf{77.7$\pm$3.5} & 82.0$\pm$3.3 & \textbf{82.7$\pm$3.3} \\
& Easy           & 78.0$\pm$9.1 & 96.0$\pm$5.8 & 96.0$\pm$5.8 & \textbf{98.0$\pm$5.0} & 98.0$\pm$5.0 & \textbf{99.0$\pm$4.4} \\
& Med.           & 30.0$\pm$6.7 & 59.5$\pm$6.9 & 70.0$\pm$6.7 & \textbf{74.5$\pm$6.5} & \textbf{79.5$\pm$6.1} & 79.0$\pm$6.2 \\
& Hard           & 37.0$\pm$5.6 & 63.3$\pm$5.6 & 72.3$\pm$5.3 & \textbf{73.0$\pm$5.3} & 78.3$\pm$5.0 & \textbf{79.7$\pm$4.9} \\
\midrule
\multirow{5}{*}{Sliding}
& Time           & $1.00{\times}$ & $2.81{\times}$ & $4.27{\times}$ & $4.11{\times}$ & $6.35{\times}$ & $5.67{\times}$ \\
& Overall        & 39.5$\pm$4.0 & 67.5$\pm$3.8 & 74.7$\pm$3.7 & \textbf{80.8$\pm$3.3} & 82.2$\pm$3.3 & \textbf{86.5$\pm$3.0} \\
& Easy           & 23.0$\pm$9.2 & 57.0$\pm$9.8 & 65.0$\pm$9.7 & \textbf{69.0$\pm$9.6} & \textbf{77.0$\pm$9.2} & 75.0$\pm$9.3 \\
& Med.           & 45.0$\pm$6.9 & 75.5$\pm$6.4 & 80.5$\pm$6.0 & \textbf{86.0$\pm$5.5} & 86.0$\pm$5.5 & \textbf{89.0$\pm$5.1} \\
& Hard           & 41.3$\pm$5.7 & 65.7$\pm$5.6 & 74.0$\pm$5.2 & \textbf{81.3$\pm$4.8} & 81.3$\pm$4.8 & \textbf{88.7$\pm$4.1} \\
\midrule
\multirow{6}{*}{\shortstack[l]{Keys and\\Doors}}
& Time (ID/OOD)  & 1.00/1.00 & 2.10/4.95 & 2.51/9.89 & 2.59/9.31 & 2.86/19.66 & 2.60/18.13 \\
& Overall        & 52.3$\pm$4.0 & 89.2$\pm$2.8 & 94.5$\pm$2.1 & \textbf{96.5$\pm$1.8} & 97.7$\pm$1.6 & \textbf{99.2$\pm$1.1} \\
& Easy           & 72.0$\pm$9.5 & 94.0$\pm$6.5 & 96.0$\pm$5.8 & \textbf{99.0$\pm$4.4} & 99.0$\pm$4.4 & \textbf{100.0$_{-3.7}$} \\
& Med.           & 48.5$\pm$6.9 & 85.5$\pm$5.5 & 90.5$\pm$4.9 & \textbf{94.5$\pm$4.1} & 97.0$\pm$3.4 & \textbf{99.0$\pm$2.6} \\
& Hard (ID)      & 48.3$\pm$5.7 & 90.0$\pm$3.9 & 96.7$\pm$2.7 & \textbf{97.0$\pm$2.6} & 97.7$\pm$2.4 & \textbf{99.0$\pm$1.9} \\
& Strict OOD     & 0.0$_{+1.3}$ & 0.0$_{+1.3}$ & 0.3$_{-0.3}^{+1.6}$ & \textbf{8.3$\pm$3.7} & 0.7$_{-0.6}^{+1.7}$ & \textbf{22.7$\pm$5.0} \\
\bottomrule
\end{tabular}
\end{table*}

\begin{table}[t]
\centering
\small
\setlength{\tabcolsep}{6pt}
\renewcommand{\arraystretch}{1.15}
\caption{\textbf{Continuous navigation on Target-Bench.} TBS improves all five trajectory metrics at both matched budgets, with the largest gains on miss rate (MR) and approach consistency (AC). \textit{Overall} is the official weighted score.}
\label{tab:targetbench_metrics}
\begin{tabular}{l ccccc c}
\toprule
\textbf{Method} & \textbf{ADE}\,$\downarrow$ & \textbf{FDE}\,$\downarrow$ & \textbf{MR\,(\%)}\,$\downarrow$ & \textbf{SE}\,$\uparrow$ & \textbf{AC}\,$\uparrow$ & \textbf{Overall}\,$\uparrow$ \\
\midrule
BoN-1 & 0.541 & 1.074 & 16.70 & 0.268 & 0.730 & 0.323 \\
\midrule
BoN-3 & 0.300 & 0.765 & 4.05  & 0.407 & 0.817 & 0.424 \\
TBS-3 & \textbf{0.276} & \textbf{0.640} & \textbf{1.73}  & \textbf{0.496} & \textbf{0.936} & \textbf{0.508} \\
\midrule
BoN-5 & 0.265 & 0.680 & 2.70  & 0.466 & 0.872 & 0.473 \\
TBS-5 & \textbf{0.263} & \textbf{0.622} & \textbf{1.56}  & \textbf{0.508} & \textbf{0.936} & \textbf{0.511} \\
\bottomrule
\end{tabular}
\vskip -.1cm
\end{table}

\begin{figure}[t]
\vskip -.5cm
\centering
\includegraphics[width=\textwidth]{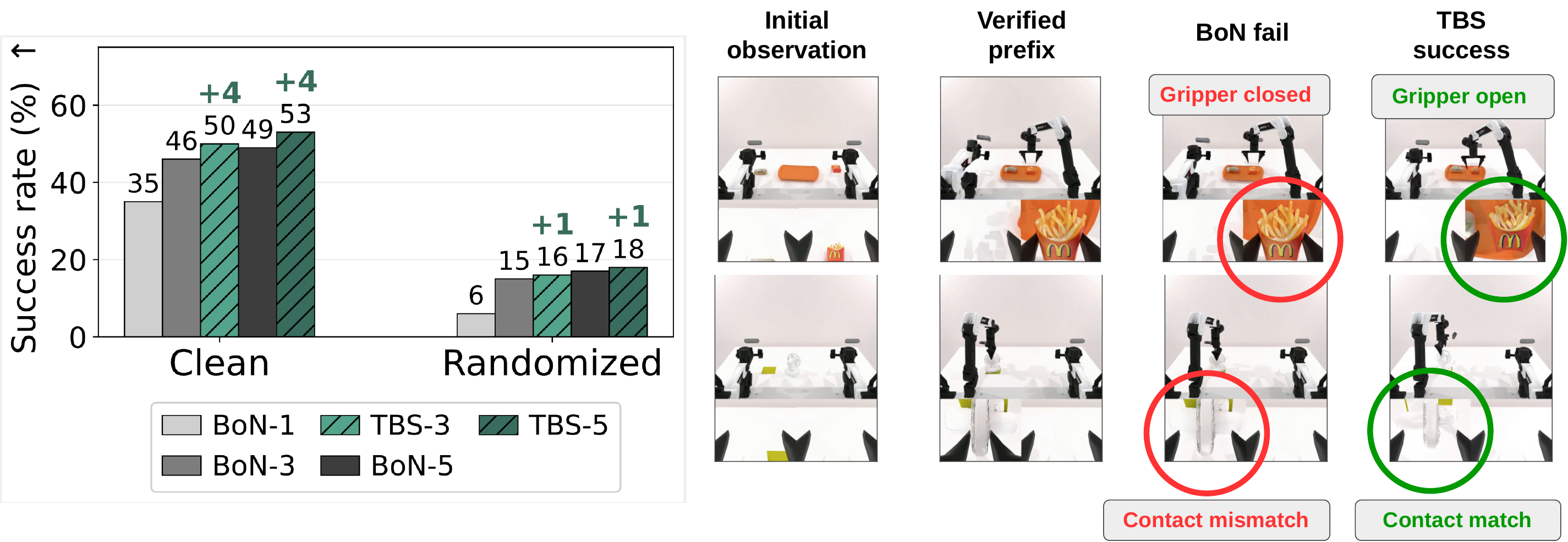}
\caption{\textbf{RoboTwin Results.} (a)~Matched-budget success rates: TBS adds $+4\%$ at every budget in clean scenes but only $+1\%$ under visual randomization. (b)~Repair sequence: TBS reuses the verified pre-grasp prefix from a failed root attempt and regenerates only the final manipulation window.}
\label{fig:robotwin_combined}
\end{figure}

\noindent\textbf{Structured reasoning: the gap widens with horizon length.}
Across all five rule-based domains (Table~\ref{tab:main_structured_results_pm}), TBS matches or exceeds matched-budget BoN on every \textit{Overall} score. The two domains with the longest path distributions show the largest improvements: comparing TBS-20 against BoN-20, Hard-split EM rises from $60.0\%$ to $81.0\%$ on 2D Maze and from $81.3\%$ to $88.7\%$ on Sliding Puzzle. On a few short-horizon splits (e.g., Sokoban Med., Sliding Easy), TBS-20 and BoN-20 fall within each other's confidence intervals, consistent with TBS targeting long-horizon failures rather than uniformly improving every problem. The clearest separation is on Keys-and-Doors \textit{Strict-OOD}: BoN-20 reaches only $0.7\%$ EM while TBS-20 reaches $22.7\%$ ($+22.0$ absolute points), a gap that Section~\ref{sec:mechanism} traces to restarted child branches. Appendix Table~\ref{tab:root_vs_restart} verifies the mechanism directly: on this split, $0/68$ TBS solves were found by depth-0 root rollouts, while all $68/68$ solved episodes came from depth-$\ge 1$ restarted branches.

\noindent\textbf{Continuous navigation: TBS improves all trajectory metrics under noisy supervision.}
Target-Bench (Table~\ref{tab:targetbench_metrics}) tests whether TBS still helps when the verifier is a learned and noisy trajectory predictor. TBS improves all five spatial metrics at both matched budgets, with the largest gains on trajectory precision: at budget $5$, TBS reduces Miss Rate from $2.70\%$ to $1.56\%$ and raises Approach Consistency from $0.872$ to $0.936$. This suggests that even a noisy process signal can localize where the rollout drifts.

\noindent\textbf{Robotic manipulation: prefix reuse helps in a physics simulator.}
On RoboTwin's clean scenes (Figure~\ref{fig:robotwin_combined}a), matched-budget TBS improves the simulator-evaluated success rate from $46\%$ (BoN-3) to $50\%$ (TBS-3), and from $49\%$ (BoN-5) to $53\%$ (TBS-5). The consistent $+4\%$ gain across both budgets indicates that the same prefix-reuse mechanism that helps on grid-worlds and continuous navigation also transfers to a closed-loop physics evaluator. We return to the visually randomized OOD setting in Section~\ref{sec:envelope}.

\begin{figure}[t]
\centering
\vskip -.2cm
\includegraphics[width=\textwidth,trim=10 30 10 30,clip,keepaspectratio]{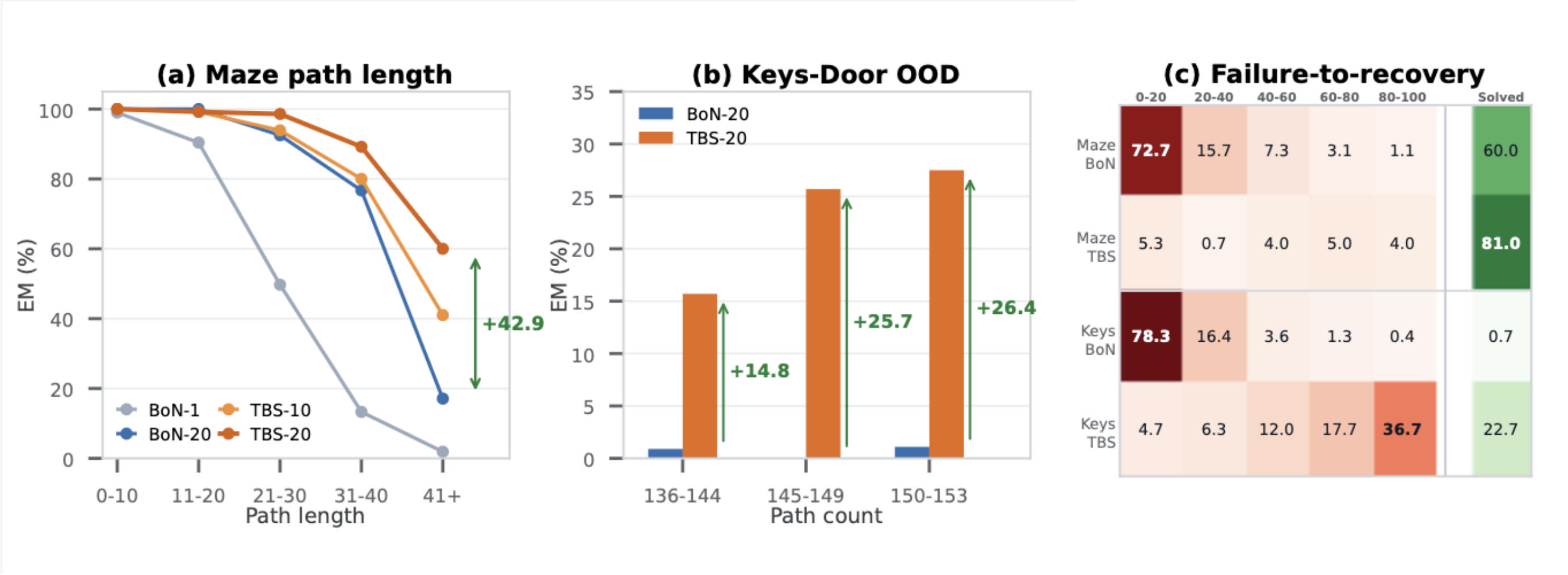}
\caption{\textbf{BoN degrades on long horizons; failures cluster at the start of trajectories.}
\textbf{(a)~2D Maze:} TBS sustains EM as path length grows, while BoN degrades sharply past $40$ steps.
\textbf{(b)~Keys-and-Doors (OOD):} TBS extends to longer paths than seen during training; BoN remains near zero.
\textbf{(c)~Failure-to-recovery distribution:} BoN failures concentrate in the first $0$--$20\%$ of trajectories, where independent root sampling repeatedly re-fails.}
\label{fig:horizon_clustering}
\vskip -.3cm
\end{figure}

\subsection{Mechanism: Prefix Reuse Targets Early Failure Bottlenecks}
\label{sec:mechanism}

Across all three regimes, rollout failures concentrate in a few short windows along the trajectory; the rest of the trajectory is locally easy (Figure~\ref{fig:horizon_clustering}). Independent root sampling has to replay these windows for every new candidate; prefix reuse keeps the verified portion and spends the budget on the residual suffix.

\noindent\textbf{Failures cluster early in structured domains.}
On 2D Maze, $72.7\%$ of failed BoN-20 rollouts have their first failure in the first $0$--$20\%$ of the trajectory; on Keys-and-Doors OOD this rises to $78.3\%$ (Figure~\ref{fig:horizon_clustering}c). Each fresh root has to replay the same early region, so BoN spends most of its budget re-failing at the same bottleneck. Once paths exceed $40$ steps, BoN-20 EM drops to $17.1\%$ while TBS-20 stays at $60.0\%$ (Figure~\ref{fig:horizon_clustering}a). The same pattern explains the strict-OOD result in Section~\ref{sec:main_results}: every TBS-20 solve came from a restarted child branch (Appendix~\ref{app:root_child}).

\noindent\textbf{Restart anchors preserve verified progress in manipulation.}
RoboTwin shows the same idea via a different anchor. When the base video-to-action policy approaches the target correctly but fails the final grasp, TBS uses the gripper-opening boundary as the restart anchor (Figure~\ref{fig:robotwin_combined}b). This preserves the pre-grasp motion (often several seconds of video) and concentrates the additional compute on regenerating the failed manipulation window.

\subsection{Ablation Studies}
\label{sec:ablations}

We ablate three design choices that drive the gains in Section~\ref{sec:main_results}: the variable-$K$ prefix conditioning interface, the depth-versus-width allocation of the search budget, and the per-domain verifier signal. Prefix conditioning and verifier quality are each individually necessary, while the depth/width allocation must be tuned per domain. Appendix~\ref{app:search_policy_ablation} additionally compares the TBS frontier against Greedy, Beam-2, and Random prefix-reuse policies under the same matched search shapes.

\noindent\textbf{Variable-$K$ prefix conditioning is required, not just nice-to-have.}
TBS resumes generation from a verified prefix by encoding it as clean prefix latents that the diffusion model conditions on. A natural alternative is to discard the prefix and restart from a single frame ($K{=}1$) at the verifier's restart anchor. We construct $40$ repairable parent prefixes per domain, each of which led to a solved child continuation in a prior TBS run, and re-present the same prefix to both interfaces under matched generation budget. Repair collapses from $100\%$ to $2.5\%$ on Maze and from $100\%$ to $5.0\%$ on Sliding when the prefix is replaced by a single-frame state reset (Appendix~\ref{app:prefix_repair}), showing that the motion content carried by the verified prefix is not redundant with the visual content of its last frame.

\noindent\textbf{Search shape adapts to the failure mode.}
At a fixed budget of $B{=}20$, we sweep the search shape $(k_1, s, m)$ and report the best EM per root width $k_1$ (Appendix~\ref{app:depth_width}). 2D Maze is depth-friendly: narrow-deep search ($k_1{=}2$) reaches $90.3\%$ and wider variants fall within $0.5$\,pp, because failures are late spatial drift on a structurally correct prefix that restart can repair. Sokoban is width-friendly: $k_1{=}2$ stalls at $76.5\%$ while $k_1{=}8$ reaches $82.7\%$ ($+6.2$\,pp), because Sokoban prefixes can be visually plausible but already strategically dead, so deeper restart from a single noisy root inherits the corruption. The headline TBS-20 rows in Table~\ref{tab:main_structured_results_pm} use the best shape per domain identified by this sweep, suggesting a simple rule of thumb: prefer depth when failures are localized at the suffix, and width when failures contaminate the prefix.

\noindent\textbf{Verifier quality matters in the harder structured domains.}
On Sokoban and Sliding Puzzle, the verifiers expose more than just prefix length. The Sokoban verifier reports BFS dead-state detection and remaining-moves count, so the priority can rank a strategically dead prefix below a shorter but live one. The Sliding Puzzle verifier flags any move whose optimal continuation is to undo it, treating the move itself as the failure rather than waiting for a downstream syntactic error. At fixed search shape, the BFS-aware Sokoban verifier improves TBS by $+4.2$ to $+10.2$\,pp over a prefix-only verifier; the undo-aware Sliding verifier improves TBS by $+2.2$ to $+4.3$\,pp (full sweeps in Appendix~\ref{app:verifier_quality_ablation}). Neither signal helps BoN at matched budget---BoN never restarts, so a better restart anchor has nothing to act on---and the asymmetry rules out a generic ``better verifier'' explanation: the gain comes from where TBS uses the verifier, not from the verifier alone.

\subsection{Operating Envelope: When Prefix Reuse Helps and When It Stops}
\label{sec:envelope}

\begin{figure}[t]
\vskip -.5cm
\centering
\includegraphics[width=\textwidth,trim=0 0 0 30,clip]{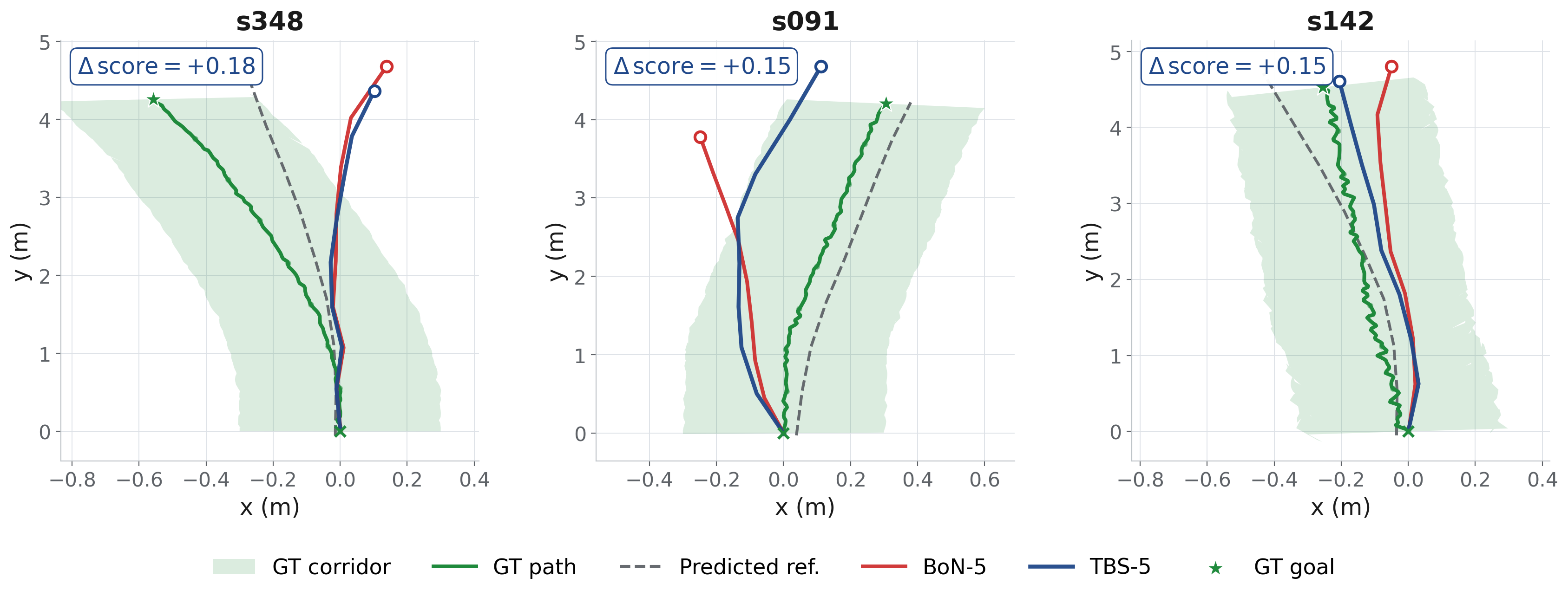}
\caption{\textbf{Trajectory refinement via temporal branching.} BoN re-ranks completed roots only (red); TBS extracts a verified prefix and resumes generation before the rollout drifts from the predicted corridor (blue), staying closer to the GT path.}
\label{fig:targetbench_overlay}
\vskip -.1cm
\end{figure}

\begin{figure}[t]
\vskip -.1cm
\centering
\includegraphics[width=0.85\textwidth]{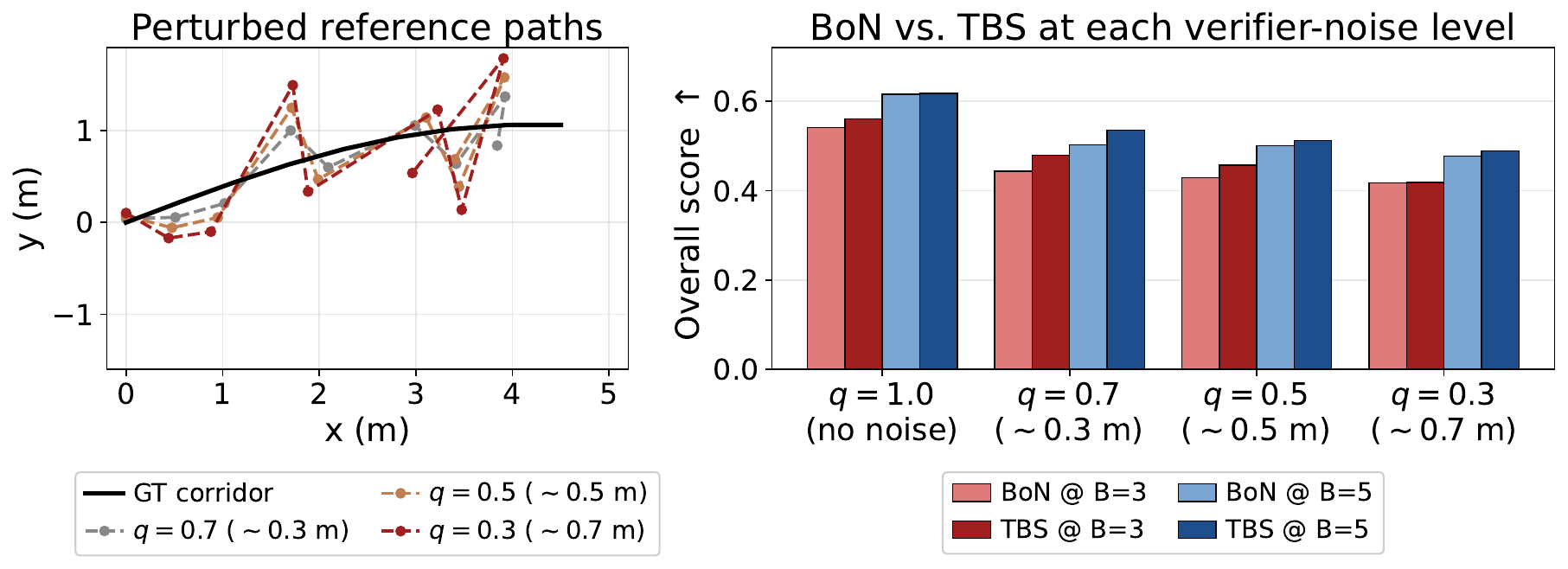}
\caption{\textbf{Robustness to verifier noise.} Target-Bench overall scores under increasing Gaussian perturbation $q$ to the reference path. TBS matches or exceeds BoN at every noise level, and converges to BoN under the strongest perturbation.}
\label{fig:targetbench_noise}
\vskip -.1cm
\end{figure}

The mechanism in Section~\ref{sec:mechanism} depends on two conditions: (i) a verifier that approximately localizes \emph{where} a trajectory deviates, and (ii) a base model that produces a repairable prefix at all. We test each condition on the regime that strains it most.

\noindent\textbf{TBS tolerates noisy process supervision.}
The Target-Bench verifier is the only one in our setup whose underlying signal is itself noisy, so we use it to stress-test condition (i). Figure~\ref{fig:targetbench_overlay} shows the qualitative behavior: TBS interrupts a rollout the moment it leaves the predicted corridor, before the drift becomes irrecoverable. To quantify this, we replace the predictor's output with a ground-truth path corrupted by Gaussian noise of increasing magnitude $q \in \{1.0, 0.7, 0.5, 0.3\}$, yielding mean ADE perturbations of approximately $0$, $0.3$, $0.5$, and $0.7$\,m. As shown in Figure~\ref{fig:targetbench_noise}, TBS matches or exceeds BoN at every noise level. Under the strongest perturbation, TBS and BoN converge at budget $3$ ($\Delta = +0.001$ overall score), and at budget $5$ TBS is still ahead by $\Delta = +0.011$. The mechanism does not need a flawless verifier; it falls back to standard root sampling as the signal weakens.

\noindent\textbf{TBS cannot supply competence the base model lacks.}
Condition (ii) is tested on RoboTwin's visually randomized OOD setting. Under heavy visual randomization the base video-to-action policy rarely reaches the target, so TBS has little to repair: BoN-3 succeeds on $15\%$ of episodes and TBS-3 on $16\%$ (Figure~\ref{fig:robotwin_combined}a). When the base policy works, TBS adds $+4\%$ to BoN at every budget; when it does not, the gain shrinks to $+1\%$. Prefix reuse amplifies competence the base model already has; it does not supply competence the base model fails to produce.

In summary, across the three regimes, TBS's gains grow with horizon length (Section~\ref{sec:mechanism}) and persist under verifier noise; but they inherit the limits of the underlying generator. Prefix reuse moves compute past the bottlenecks where the generator predictably fails; it does not change what the generator can locally produce.

\section{Discussion}
\label{sec:discussion}

This work shows that single-shot video generation bottlenecks long-horizon reasoning, and proposes Temporal Backtracking Search (TBS) to shift test-time scaling to the temporal axis. By caching verified partial progress and reallocating compute to repair localized errors, TBS unlocks stronger reasoning across symbolic, learned, and physics-based environments.

\textbf{Limitations and Future Work.} While TBS significantly improves long-horizon performance, it still depends on the base video model's local competence: the model must generate at least a short valid prefix before search can repair later errors, which may fail under extreme OOD visual conditions. Our current implementation also relies on domain-specific external verifiers; future work could internalize this supervision by training implicit video process reward models (PRMs), enabling autonomous temporal search across broader domains without external simulators.

\begin{ack}
    This work was supported in part by a gift from the Chan Zuckerberg Initiative Foundation to establish the Kempner Institute at Harvard University. We also thank Pickle Robotics and AWS AGI Labs for their support of the lab. 
\end{ack}

\bibliographystyle{plainnat}
\bibliography{references}

\newpage
\appendix       
\etocdepthtag.toc{appendix}
\etocsettagdepth{main}{none} 
\etocsettagdepth{appendix}{subsection} 
\etocsettocstyle{\section*{Appendix Contents}}{}
\tableofcontents
\vspace{1em}
\section{Terminology and Claim Boundaries}
\label{app:claim_boundaries}
\providecommand{\NEW}[1]{{\color{magenta}[NEW: #1]}}

\paragraph{Search policy.}
TBS is implemented as prefix-priority best-first search over a global frontier of verified prefixes. It is not UCT-style Monte Carlo tree search. The paper claims that prefix reuse with frontier search improves long-horizon generative video reasoning. It does not claim that the current search policy is a full symbolic planner.

\paragraph{Verifier scope.}
The structured-domain verifiers are domain-specific process verifiers. They can localize the first checked failure frame because the tasks are grid-world or grid-like rule-based problems. They do not provide the correct continuation. TBS receives a verified prefix, restart anchor, and priority score. The generator must still synthesize the suffix.

\section{Structured Video Reasoning Details}
\label{app:structured_video_reasoning}

This appendix collects the structured video-reasoning details: prefix-conditioning ablation, training settings, domain definitions, prompts, symbolic verifiers, search budgets, extended results, failure clustering, root-versus-child solves, and depth--width search behavior. It also includes verifier-quality ablations and search-policy ablations.

\subsection{Prefix-Conditioning Ablation}
\label{app:prefix_repair}

TBS presents the verified prefix to the generator as clean prefix latents that the diffusion model conditions on while denoising the regenerated suffix. We compare two policies that branch from the same verified parent prefix under matched generation budget:

\begin{itemize}
\item \texttt{full\_prefix} (TBS default): the verified prefix frames are encoded as prefix latents with their actual motion content.
\item \texttt{last\_frame\_reset}: only the last verified frame is encoded as a single $K{=}1$ first-frame condition. Prefix-length conditioning and motion content are both discarded. This is the naive single-frame state reset.
\end{itemize}

\paragraph{Subset construction.}
The 40-sample evaluation subset per domain is constructed as follows. From a prior \texttt{full\_prefix} TBS run, we select samples where a child branch at depth $\ge 1$ achieved EM, then trace back to the failed parent record from which that child was branched. We retain only samples where the parent prefix is at least 5 pixel frames long at a multi-frame conditioning boundary, and take the first 40 such samples in dataset order. The same parent prefix is then re-presented to the generator under each conditioning policy at matched generation budget.

By construction, \texttt{full\_prefix} extends the parent prefix to a solved trajectory in $100\%$ of cases (the manifest is filtered to ``parent prefixes that \texttt{full\_prefix} provably repairs''). The ablation therefore does not measure population-level success rates; it measures how much of \texttt{full\_prefix}'s repair capability the simpler conditioning policy can match given the identical parent prefix and identical budget.

\begin{table}[h]
\centering\small
\setlength{\tabcolsep}{4pt}
\begin{tabular}{llcc}
\toprule
Task & Policy & EM & SR (any goal) \\
\midrule
Maze    & \texttt{last\_frame\_reset}         & $1/40 = 2.5\%$  & $2/40 = 5.0\%$ \\
Maze    & \texttt{full\_prefix} (TBS)          & $40/40 = 100\%$  & $40/40 = 100\%$ \\
\midrule
Sliding & \texttt{last\_frame\_reset}         & $2/40 = 5.0\%$  & $2/40 = 5.0\%$ \\
Sliding & \texttt{full\_prefix} (TBS)          & $40/40 = 100\%$  & $40/40 = 100\%$ \\
\bottomrule
\end{tabular}

\caption{Prefix-conditioning ablation on known-repairable parent prefixes. \texttt{full\_prefix} is $100\%$ by construction. Replacing the full prefix with a single last-frame condition collapses recovery to $2.5\%$ on Maze and $5.0\%$ on Sliding.}
\label{tab:prefix_repair_ablation}
\end{table}

\subsection{Training and Inference Details}
\label{app:training}

All TBS results use a single base video diffusion model (Wan2.2-TI2V-5B~\citep{wan2025}) fine-tuned with domain-specific LoRA adapters that support variable-$K$ prefix conditioning. The base model weights are frozen throughout; only LoRA parameters are updated. A separate LoRA adapter is trained per domain. All training is performed on pre-encoded VAE latent caches.

\subsubsection{LoRA hyperparameters}

\begin{table}[h]
\centering\small
\begin{tabular}{lc}
\toprule
Hyperparameter & Value \\
\midrule
Base model & Wan2.2-TI2V-5B \\
LoRA rank / alpha & $32 / 32$ \\
Optimizer & AdamW \\
Learning rate & $2 \times 10^{-4}$ \\
Weight decay & $1 \times 10^{-4}$ \\
Batch size / grad accumulation & $1 / 2$ \\
Epochs & 100 \\
Dataset repeat & 2 \\
Resolution & $512 \times 512$ \\
Frame count & 81 (161 for strict-OOD Keys-Doors and 24-grid 3D Maze) \\
Saved FPS & 24 \\
LR scheduler & constant (no warmup) \\
Inference denoising steps & 50 \\
CFG scale & 5.0 \\
Sigma shift & 5.0 \\
VAE compression $\tau$ & 4 (each latent frame $\equiv$ 4 pixel frames) \\
Seed schedule & base seed 0; sample seed = base + sample\_index \\
\bottomrule
\end{tabular}
\caption{LoRA fine-tuning and inference hyperparameters shared across all domains.}
\label{tab:training_hparams}
\end{table}

\subsubsection{Variable-$K$ conditioning curriculum}

The model is trained with a curriculum over prefix length $K$ that progressively increases the maximum allowed prefix and shifts probability mass toward longer prefixes. Prefix lengths are sampled from four buckets: $K{\in}\{1\}$, $K{\in}\{2{-}4\}$, $K{\in}\{5{-}8\}$, $K{\in}\{9+\}$. Without this curriculum the model fails to learn long-prefix conditioning and collapses on $K{>}5$ at inference.

\begin{table}[h]
\centering\small
\begin{tabular}{ccc}
\toprule
Training progress & Max $K$ & Bucket weights $(K{=}1,\ 2{-}4,\ 5{-}8,\ 9+)$ \\
\midrule
$0$--$10\%$    & 1  & $(1.00, 0.00, 0.00, 0.00)$ \\
$10$--$30\%$   & 5  & $(0.50, 0.30, 0.20, 0.00)$ \\
$30$--$55\%$   & 10 & $(0.40, 0.30, 0.20, 0.10)$ \\
$55$--$80\%$   & 15 & $(0.35, 0.30, 0.20, 0.15)$ \\
$80$--$100\%$  & 20 & $(0.35, 0.25, 0.20, 0.20)$ \\
\bottomrule
\end{tabular}
\caption{Variable-$K$ training curriculum.}
\label{tab:curriculum}
\end{table}

\subsubsection{Per-frame timestep masking}                                                
During training, the first $K$ latent frames receive timestep $t{=}0$ (clean signal, $\sigma_{t_i}{=}0$), and the         
remaining suffix frames receive a single sampled timestep $t$. This per-frame timestep mask teaches the model to treat the
first $K$ frames as given context and regenerate only the continuation. Equation~\ref{eq:variable_k_mask} in the main    
text gives the formal rectified-flow forward process.

\subsection{Structured Video Reasoning Domains}
\label{app:domains}

We evaluate on five rule-based domains. Each is a 2D top-down or 3D isometric rendering of a sequential decision problem, governed by deterministic rules so that an exact symbolic verifier can return both terminal success (EM) and process-level failure information. Test sets contain 600 samples per domain split as 100 easy / 200 medium / 300 hard, plus a 300-sample strict-OOD split for Keys-and-Doors.

\paragraph{2D Maze.} Grid sizes $7\!\times\!7$ (easy), $11\!\times\!11$ (medium), $15\!\times\!15$ (hard). A red agent must traverse the white-corridor maze to a green goal without entering the light-blue walls. Solution path lengths range from 2 to 78 steps.

\paragraph{3D Maze.} First-person rendered isometric mazes with stacked paths and ladders. The agent (yellow ball) must climb to a red goal cube. Path-following structure mirrors 2D Maze, but visual tracking is harder.

\paragraph{Sliding Puzzle.} Grid sizes $3\!\times\!3$ (easy), $4\!\times\!4$ (medium), $5\!\times\!5$ (hard). The agent must arrange numbered tiles in row-major order with the blank in the bottom-right. The goal arrangement is not visible in the conditioning image, so failures are semantic rather than visual wall-crossings. Tiles are rendered in large font for OCR readability; the easy split is paradoxically harder because $3\!\times\!3$ has smaller tiles.

\paragraph{Sokoban.} Board sizes $5\!\times\!5$ / $8\!\times\!8$ / $12\!\times\!12$ with wall probabilities $0.08 / 0.12 / 0.16$. Single box, single target. The blue agent can push but not pull; a visually valid prefix can already be in a dead state where no solution exists.

\paragraph{Keys-and-Doors.} The agent must collect colored keys in the correct sequence before reaching the door. Adds long-horizon sub-goal ordering on top of path-following. The strict-OOD split uses 161-frame videos with path lengths exceeding the training distribution (typically 136--153 path cells); root sampling on this split nearly collapses (BoN-20 EM $0.7\%$).

\subsubsection{Train / test splits}

\begin{table}[h]
\centering\small
\begin{tabular}{lccc}
\toprule
Domain  & Easy & Medium & Hard \\
\midrule
2D Maze         & 500 & 500 & 500 \\
3D Maze         & 500 & 500 & 500 \\
Sokoban         & 500 & 500 & 500 \\
Sliding Puzzle  & 500 & 500 & 500 \\
Keys-and-Doors  & 500 & 500 & 500 \\
\midrule
\textit{Test (all domains)} & 100 & 200 & 300 \\
\textit{Keys-OOD strict split} & \multicolumn{3}{c}{300 samples, 161-frame videos} \\
\bottomrule
\end{tabular}
\caption{Dataset sizes per domain. All domains use 1500 training samples balanced across difficulty. The Keys-and-Doors strict-OOD split uses path lengths that exceed the training distribution.}
\label{tab:datasets}
\end{table}

\subsection{Prompt Templates}
\label{app:prompts}

All rich prompts share a four-block structure: object identities and colors; movement constraints; scene invariance; camera constraints. The prompts are uniform across difficulty levels within each domain---difficulty is determined by the input image, not by the text.

\subsubsection{2D Maze}
\begin{quote}\small
Create a 2D animation based on the provided image of a maze. The red circle slides smoothly along the white square path, stopping perfectly on the green square. The red circle never slides or crosses into the light blue square areas of the maze. The camera is a static, top-down view showing the entire maze.

\textit{Maze:} The maze paths are white square, the walls are light blue square. The red circle moves to the goal position, represented by green square. The red circle slides smoothly along the white square path. The red circle stops perfectly on the green square.

\textit{Scene:} No change in scene composition. No change in the layout of the maze. The red circle travels along the white square path without speeding up or slowing down.

\textit{Camera:} Static camera. No zoom. No pan. No glitches, noise, or artifacts.
\end{quote}

\subsubsection{3D Maze}
\begin{quote}\small
Create a 3D animation based on the provided image of a cube maze. A yellow ball slides smoothly along the gray cube pathway, climbs up the vertical ladders step by step, and finally stops perfectly on the red cube at the top. The yellow ball never touches or passes through the blue cube or any non-gray areas of the maze. The camera remains static in an isometric, top-down angle showing the entire structure.

\textit{Maze:} The maze consists of stacked transparent gray cubes forming a 3D pathway. The red cube represents the goal position. The blue cube marks the starting platform where the yellow ball begins. The yellow ball moves upward along the gray path, climbing vertically via the ladders. The ball slides smoothly without sudden changes in direction or speed. The ball stops exactly on top of the red cube at the end.

\textit{Scene:} No structural or color changes during animation. The maze layout and cube arrangement remain unchanged. The yellow ball moves continuously at a constant speed along the 3D path.

\textit{Camera:} Static camera. No zoom. No pan. No glitches, noise, or artifacts.
\end{quote}

\subsubsection{Sokoban}
\begin{quote}\small
Create a 2D animation based on the provided image of a grid puzzle. The blue ball moves into position behind the yellow square and smoothly pushes it toward the red goal square. The yellow square only slides when pushed from behind by the blue ball and moves in a straight line along the white floor tiles. When the direction of the yellow square's movement needs to change, the blue ball must reposition itself to a new side of the yellow square. The yellow square never crosses or overlaps any gray walls. The camera is a static, top-down view showing the entire puzzle.

\textit{Maze:} The floor area is white, and the walls are gray. The yellow square can only move when pushed by the blue ball from behind. The blue ball cannot pull the yellow square or move through walls. The yellow square slides smoothly in one direction until it reaches the red goal square. The animation stops perfectly when the yellow square aligns with the red goal square.

\textit{Scene:} No change in scene composition. No change in the layout of the puzzle. The movement is smooth, with no speed variation.

\textit{Camera:} Static camera. No zoom. No pan. No glitches, noise, or artifacts.
\end{quote}

\subsubsection{Sliding Puzzle}
\begin{quote}\small
Complete this sliding puzzle by arranging the numbered tiles in sequential order. The goal state has tiles numbered left-to-right, top-to-bottom, with the blank space at the bottom-right corner.

\textit{Puzzle:} The grid contains numbered tiles and one blank space (white cell). A tile can only move if it is directly adjacent (up, down, left, or right) to the blank space. Each move slides exactly one adjacent tile into the blank space; the tile and blank space swap positions. Only one tile moves per step; all other tiles remain stationary during each move. The number displayed on each tile must remain the same throughout the entire animation; tiles must not change, swap, or lose their numbers at any point. The sequence of moves must follow a shortest valid solution path from the initial state to the goal state; do not make unnecessary detours or extra moves.

\textit{Scene:} No change in scene composition. No change in the grid layout, tile sizes, or tile colors. Tiles slide smoothly without speeding up or slowing down. The background and grid structure remain unchanged throughout.

\textit{Camera:} Static camera. No zoom. No pan. No glitches, noise, or artifacts.
\end{quote}

\subsubsection{Keys-and-Doors}
\begin{quote}\small
Create a 2D animation based on the provided image of a maze. The green circle is the agent. There are multiple colored diamond-shaped keys and one hollow rectangle door in the maze. The green circle slides smoothly along the white paths, using the shortest valid paths to collect all visible keys in the order that minimizes the total travel distance, then moves to the door. Each key disappears when the green circle reaches it. The green circle never crosses or overlaps any black wall areas of the maze. The camera is a static, top-down view showing the entire maze.

\textit{Maze:} The maze paths are white, and the walls are black. The green circle must collect all visible keys before reaching the door. The green circle slides smoothly along the white paths. The green circle never slides or crosses into the black wall areas. The green circle stops perfectly on the door after collecting all keys.

\textit{Scene:} No change in scene composition. No change in the layout of the maze. The movement is smooth, with no speed variation.

\textit{Camera:} Static camera. No zoom. No pan. No glitches, noise, or artifacts.
\end{quote}

\subsection{Verifier and Restart Details}
\label{app:verifier}

For every structured video-reasoning domain, the verifier returns a process trace (\texttt{first\_failure\_frame}, \texttt{restart\_frame}, \texttt{prefix\_pixel\_frames}). The verifier exposes a failure location and a restartable prefix; it does not provide the correct continuation.

\subsubsection{2D Maze and 3D Maze}
The 2D Maze verifier extracts the player sprite template from the first frame using its known starting bounding box, then locates the player at each subsequent frame via normalized cross-correlation. Pixel coordinates are mapped to grid cells using the known maze geometry, and each cell transition is checked against the maze graph, wall layout, and task-derived goal/progress constraints. The verifier does not receive or return the future solution trajectory. The first failure frame is the earliest of motion / tracking / appearance / duplicate-player / wall-crossing / length failures. The restart frame is backed up to the entry point of the last correctly visited cell, then aligned to the VAE's latent frame boundary. The 3D Maze verifier uses the same logic on first-person renderings and additional blob diagnostics for ladder identity swaps.

\subsubsection{Sokoban}
The Sokoban verifier tracks both the player and the box via template matching, maps centers to grid cells, and validates wall collisions and legal push transitions. It additionally runs a single-box BFS state check for remaining moves and dead-state detection: \texttt{bfs\_initial\_moves} (BFS shortest path from initial state), \texttt{bfs\_remaining\_moves} (BFS from current state), \texttt{bfs\_dead\_state} (true if the box is cornered and no solution exists), and \texttt{bfs\_progress} $= 1 - \texttt{remaining}/\texttt{initial}$. Sokoban TBS uses progress-aware priority (Section~\ref{app:tbs_configs}).

\subsubsection{Sliding Puzzle}
The Sliding Puzzle verifier reads tile values via OCR (CPU EasyOCR) at stable frames where the board is not mid-transition, simulates board transitions between consecutive stable states, and verifies that exactly one adjacent tile moved into the blank cell at each transition. It detects illegal multi-tile moves, duplicate tiles, wrong board states, and stalled segments.

In addition, an \emph{undo-required} semantic policy checks at every stable state whether the only optimal next move is to undo the previous move. If yes, the previous move was a strategic mistake and the verifier triggers a rollback to the prior stable state. This cuts prefixes at semantic dead-ends rather than at superficial OCR failures, providing cleaner restart anchors for search.

\subsubsection{Keys-and-Doors}
Extends the maze verifier with key-order checking. At each key-pickup position the verifier confirms the player collected the key, and after all keys are collected confirms the door is reached. OOD evaluation uses 161-frame videos with path lengths exceeding the training distribution.

\subsection{TBS Search Budgets and Priority}
\label{app:tbs_configs}

\paragraph{Budget formula.}
A TBS configuration is parameterized by root width $k_1$, samples per expanded non-root node $s$, and maximum expansions $m$. Total generation budget is
\begin{equation}
B = k_1 + s\,(m-1).
\label{eq:appendix_budget}
\end{equation}
Matched-budget BoN samples $N{=}B$ independent root rollouts.

\paragraph{Domain-specific priority exceptions.}
\begin{itemize}
\item \textbf{Sokoban (progress-aware)} The frontier priority ranks first by EM, then by BFS state quality (non-dead state, fewer remaining moves, higher progress), and only falls back to prefix length. The BFS state-quality signal therefore takes precedence over raw prefix length.

\item \textbf{Sliding Puzzle.} The undo-required policy can roll the restart anchor back to the previous stable board state when the current move creates a semantic dead-end.
\end{itemize}

\paragraph{Per-row TBS configurations.}
Each row of Table~\ref{tab:main_structured_results_pm} in the main paper corresponds to a specific $(k_1, s, m)$ tuple:

\begin{table}[h]
\centering\small
\begin{tabular}{llcccl}
\toprule
Domain & Budget & $k_1$ & $s$ & $m$ & Priority \\
\midrule
Maze            & TBS-10 & 2 & 2 & 5 & default prefix priority \\
Maze            & TBS-20 & 2 & 2 & 10 & default prefix priority \\
Maze3D          & TBS-10 & 8 & 2 & 2 & default \\
Maze3D          & TBS-20 & 8 & 2 & 7 & default \\
Sokoban         & TBS-10 & 8 & 2 & 2 & progress-aware (BFS) \\
Sokoban         & TBS-20 & 8 & 2 & 7 & progress-aware (BFS) \\
Sliding Puzzle  & TBS-10 & 6 & 2 & 3 & default + undo-required policy \\
Sliding Puzzle  & TBS-20 & 5 & 3 & 6 & default + undo-required policy \\
Keys-Doors ID   & TBS-10 & 6 & 2 & 3 & default \\
Keys-Doors ID   & TBS-20 & 4 & 2 & 9 & default \\
Keys-Doors OOD  & TBS-10 & 2 & 2 & 5 & default \\
Keys-Doors OOD  & TBS-20 & 4 & 2 & 9 & default \\
\bottomrule
\end{tabular}
\caption{Structured-domain TBS search configurations. Domain-specific shapes were chosen via the depth-versus-width sweep in Section~\ref{app:depth_width}. Total generation budget $B = k_1 + s(m-1)$.}
\label{tab:tbs_configs}
\end{table}

\subsection{Structured Video Reasoning Extended Results}
\label{app:structured_extra}

\subsubsection{Confidence interval methodology}

We report 95\% Wilson confidence half-widths because EM is a binomial proportion: a sample either solves a problem or does not. Wilson is preferred over Wald because Wald collapses at $0\%$ and $100\%$. For a proportion $p = k/n$ with $z = 1.959964$:
\begin{align}
c &= \frac{p + z^2/(2n)}{1 + z^2/n}, \quad
h = \frac{z\sqrt{p(1{-}p)/n + z^2/(4n^2)}}{1 + z^2/n}, \notag\\
\text{CI}_{95} &= [\,\max(0, c - h),\ \min(1, c + h)\,].
\label{eq:wilson}
\end{align}
The displayed $\pm$ is the larger of the upper and lower halves, so the interval contains the full Wilson CI even at saturation. Sample sizes are $n{=}600$ overall, $n{=}100$ easy, $n{=}200$ medium, and $n{=}300$ hard or strict-OOD.

\subsubsection{Maze rich-prompt path-length breakdown}

\begin{table}[h]
\centering\small
\begin{tabular}{lcccccc}
\toprule
Method & 0--10 & 11--20 & 21--30 & 31--40 & 41+ & Total \\
\midrule
BoN-1   &  99.0 &  90.4 & 49.7 & 13.3 &  1.9 & 51.0 \\
BoN-5   & 100.0 &  99.2 & 78.9 & 47.5 &  5.7 & 67.7 \\
BoN-10  & 100.0 & 100.0 & 89.8 & 65.0 & 12.4 & 75.2 \\
BoN-20  & 100.0 & 100.0 & 92.5 & 76.7 & 17.1 & 79.0 \\
TBS-10  & 100.0 &  99.2 & 93.9 & 80.0 & 41.0 & 84.0 \\
TBS-20  & 100.0 &  99.2 & 98.6 & 89.2 & 60.0 & 90.3 \\
\bottomrule
\end{tabular}
\caption{2D Maze rich-prompt EM by shortest-path-length bin. The bins are used only for post-hoc analysis of horizon length, not as an online solution signal for TBS. The TBS gain grows monotonically with horizon: at $41+$ steps, BoN-20 collapses to $17.1\%$ while TBS-20 reaches $60.0\%$ ($+42.9$\,pp). This is the numerical table backing Figure~\ref{fig:horizon_clustering}(a).}
\label{tab:maze_pathlen}
\end{table}

\subsubsection{Keys-and-Doors strict-OOD path-cell breakdown}

\begin{table}[h]
\centering\small
\begin{tabular}{lcccc}
\toprule
& \multicolumn{3}{c}{Path cells (strict-OOD)} & \\
\cmidrule(lr){2-4}
Method & 136--144 ($n{=}108$) & 145--149 ($n{=}101$) & 150--153 ($n{=}91$) & Total \\
\midrule
BoN-20  &  0.9 &  0.0 &  1.1 &  0.7 \\
TBS-20  & 15.7 & 25.7 & 27.5 & 22.7 \\
$\Delta$ & $+14.8$ & $+25.7$ & $+26.4$ & $+22.0$ \\
\bottomrule
\end{tabular}
\caption{Keys-and-Doors strict-OOD EM by path-cell-count bin. BoN-20 stays below $1.1\%$ across all bins; TBS-20 reaches $15.7$--$27.5\%$. Numerical version of Figure~\ref{fig:horizon_clustering}(b).}
\label{tab:keys_ood_pathlen}
\end{table}

\subsection{Failure Clustering Statistics}
\label{app:failure_clustering}

This section provides the numerical values behind the failure-to-recovery panel of Figure~\ref{fig:horizon_clustering}(c) and quantifies the clustering observation that motivates TBS in Section~1.

\subsubsection{Failure-to-recovery distribution}

\begin{table}[h]
\centering\small
\setlength{\tabcolsep}{4pt}
\begin{tabular}{llrrrrrr}
\toprule
Domain & Row & 0--20\% & 20--40\% & 40--60\% & 60--80\% & 80--100\% & Solved \\
\midrule
2D Maze   & BoN-20 failures   & 72.7 & 15.7 &  7.3 &  3.1 & 1.1 &   0.0 \\
2D Maze   & TBS-20 outcomes   &  5.3 &  0.7 &  4.0 &  5.0 & 4.0 &  81.0 \\
\addlinespace[2pt]
3D Maze   & BoN-20 failures   & 21.6 & 31.9 & 27.5 & 14.7 & 4.4 &   0.0 \\
3D Maze   & TBS-20 outcomes   &  2.0 &  0.7 &  1.0 &  1.0 & 0.0 &  95.3 \\
\addlinespace[2pt]
Sokoban   & BoN-20 failures   &  8.3 & 18.6 & 22.7 & 22.2 &28.1 &   0.0 \\
Sokoban   & TBS-20 outcomes   &  0.3 &  2.3 &  2.0 &  8.7 & 7.0 &  79.7 \\
\addlinespace[2pt]
Sliding Puzzle   & BoN-20 failures   & 25.7 & 35.2 & 17.2 & 16.1 & 5.7 &   0.0 \\
Sliding Puzzle   & TBS-20 outcomes   &  2.0 &  0.7 &  2.3 &  3.7 & 2.7 &  88.7 \\
\addlinespace[2pt]
Keys-and-Doors ID   & BoN-20 failures   & 14.3 & 38.8 & 30.8 & 12.8 & 3.3 &   0.0 \\
Keys-and-Doors ID   & TBS-20 outcomes   &  0.0 &  0.0 &  0.3 &  0.3 & 0.3 &  99.0 \\
\addlinespace[2pt]
Keys-and-Doors-OOD  & BoN-20 failures   & 78.3 & 16.4 &  3.6 &  1.3 & 0.4 &   0.0 \\
Keys-and-Doors-OOD  & TBS-20 outcomes   &  4.7 &  6.3 & 12.0 & 17.7 &36.7 &  22.7 \\
\bottomrule
\end{tabular}
\caption{Failure-to-recovery distribution by trajectory-progress bin. ``BoN-20 failures'' rows give the distribution of failed BoN-20 root samples by their farthest verified-progress bin (sums to 100\%). ``TBS-20 outcomes'' rows give the distribution of TBS-20 final episode outcomes; the right-most column is TBS hard-split EM. Maze and Keys-and-Doors-OOD show the strongest clustering: more than $70\%$ of BoN failures concentrate in the earliest $0$--$20\%$ bin, exactly the region where independent root sampling re-fails.}
\label{tab:fail_to_recovery}
\end{table}

\subsubsection{Densest local-window share}

\begin{table}[h]
\centering
\scriptsize
\begin{tabular}{lrrrrl}
\toprule
Domain & Failed records & Median densest share & $\ge 50\%$ in window & $\ge 70\%$ in window & Dominant window \\
\midrule
2D Maze            & 4920 & 77.8\% & 92.9\% & 64.5\% & frames 1--16 \\
3D Maze            & 1522 & 93.9\% &100.0\% & 86.8\% & frames 19.5--34.5 \\
Sokoban            & 3830 & 57.1\% & 71.9\% & 19.7\% & frames 41--56 \\
Sliding Puzzle     & 4306 & 66.7\% & 88.7\% & 44.9\% & frames 14--29 \\
Keys-and-Doors ID  & 2918 & 71.4\% & 93.3\% & 53.7\% & frames 21--36 \\
Keys-Doors strict-OOD & 5997 & 85.0\% & 98.7\% & 81.3\% & frames 1--32 \\
\bottomrule
\end{tabular}
\caption{For every hard problem we compute the median fraction of failed BoN-20 seeds that fall within a single 16-frame local window (32 frames for the 161-frame strict-OOD videos, so the window covers the same fraction of the trajectory). Local clustering is strong everywhere except Sokoban.}
\label{tab:dense_window}
\end{table}

\subsection{Root Solves vs.\ Restarted-Branch Solves}
\label{app:root_child}

This section directly probes the TBS mechanism: of the episodes solved by TBS, how many were already solved by a root rollout (depth $0$) and how many required a restarted child branch (depth $\ge 1$). If TBS gains came purely from extra root sampling, the depth-0 share would be high; if they came from prefix reuse, the depth-$\ge 1$ share dominates.

\begin{table}[h]
\centering\footnotesize
\begin{tabular}{lrrrr}
\toprule
Domain & TBS-20 final & Root solves (d=0) & Restart solves (d$\ge$1) & Restart-added EM \\
\midrule
2D Maze            & $542/600 = 90.3\%$ & 357 & 185 & $+30.8$\,pp \\
3D Maze            & $586/600 = 97.7\%$ & 574 & 12 & $+2.0$\,pp \\
Sokoban            & $496/600 = 82.7\%$ & 448 & 48 & $+8.0$\,pp \\
Sliding Puzzle     & $519/600 = 86.5\%$ & 405 & 114 & $+19.0$\,pp \\
Keys-and-Doors ID  & $595/600 = 99.2\%$ & 510 & 85 & $+14.2$\,pp \\
Keys-and-Doors-OOD & $68/300 = 22.7\%$ & 0 & 68 & $+22.7$\,pp \\
\bottomrule
\end{tabular}
\caption{Depth at which each TBS-20 solve was first achieved. Root solves are solves found at depth 0 inside the TBS run; a child branch first solves restarted solves. Keys-and-Doors-OOD is the cleanest mechanism case: every solved episode comes from a restarted branch.}
\label{tab:root_vs_restart}
\end{table}

\begin{table}[h]
\centering\small
\begin{tabular}{lrrrr}
\toprule
Domain & Depth 0 & Depth 1 & Depth 2 & Depth 3+ \\
\midrule
2D Maze     & $357/542 = 65.9\%$ & $70/542 = 12.9\%$ & $50/542 = 9.2\%$ & $65/542 = 12.0\%$ \\
3D Maze     & $574/586 = 98.0\%$ & $9/586 = 1.5\%$ & $2/586 = 0.3\%$ & $1/586 = 0.2\%$ \\
Sokoban     & $448/496 = 90.3\%$ & $32/496 = 6.5\%$ & $12/496 = 2.4\%$ & $4/496 = 0.8\%$ \\
Sliding     & $405/519 = 78.0\%$ & $76/519 = 14.6\%$ & $24/519 = 4.6\%$ & $14/519 = 2.7\%$ \\
Keys ID     & $510/595 = 85.7\%$ & $63/595 = 10.6\%$ & $18/595 = 3.0\%$ & $4/595 = 0.7\%$ \\
Keys-OOD    & $0/68 = 0.0\%$ & $11/68 = 16.2\%$ & $16/68 = 23.5\%$ & $41/68 = 60.3\%$ \\
\bottomrule
\end{tabular}
\caption{First-solved depth distribution among solved TBS-20 episodes. Depth 0 denotes a root solve; depth $\ge 1$ denotes at least one temporal restart.}
\label{tab:depth_solve_rate}
\end{table}

\paragraph{Cumulative EM by search depth (2D Maze).}
On Maze, TBS continues to gain across multiple restart depths rather than saturating after the first restart.

\begin{table}[h]
\centering\small
\begin{tabular}{rrrrr}
\toprule
Depth & Reached & New solves at depth & Solve rate when reached & Cumulative EM \\
\midrule
0 & 600 & 357 & 59.5\% & 59.5\% \\
1 & 243 &  70 & 28.8\% & 71.2\% \\
2 & 173 &  50 & 28.9\% & 79.5\% \\
3 & 120 &  33 & 27.5\% & 85.0\% \\
4 &  83 &  15 & 18.1\% & 87.5\% \\
5 &  55 &  14 & 25.5\% & 89.8\% \\
6 &  27 &   3 & 11.1\% & 90.3\% \\
\bottomrule
\end{tabular}
\caption{2D Maze TBS depth-by-depth. ``Reached'' is the number of problems where TBS expanded a node at this depth; ``new solves at depth'' is the number of problems first solved at exactly this depth. The solve rate stays above $25\%$ through depth 3, evidence that reused prefixes remain structurally useful after multiple continuations.}
\label{tab:cumulative_depth_maze}
\end{table}

\subsection{Root-Width Sensitivity at Matched Budget}
\label{app:root_sensitivity}

This subsection reports the \emph{best EM per domain} as $k_1$ varies at the headline budget $B{=}20$, holding the total generation count fixed. For each $k_1$ value, we take the maximum EM across all $(s, m)$ combinations satisfying $k_1 + s(m{-}1) = 20$ in the per-domain sweep. The range across $k_1$ measures how sensitive each domain is to the choice of root width.

\begin{table}[h]
\centering\small
\setlength{\tabcolsep}{4pt}
\begin{tabular}{lrrrrr}
\toprule
Domain & $k_1{=}2$ & $k_1{=}4$ & $k_1{=}5$ & $k_1{=}8$ & Range \\
\midrule
2D Maze              & 90.3\% & 89.8\% & 89.8\% & 89.8\% & 0.5\,pp \\
3D Maze              & 97.0\% & 97.3\% & 97.3\% & 97.7\% & 0.7\,pp \\
Sokoban              & 76.5\% & 81.2\% & 82.0\% & \textbf{82.7\%} & \textbf{6.2\,pp} \\
Sliding              & 85.0\% & 85.8\% & 86.5\% & 86.5\% & 1.5\,pp \\
Keys-Doors ID        & 97.3\% & \textbf{99.2\%} & 99.0\% & 99.0\% & 1.9\,pp \\
Keys-Doors strict-OOD & 21.3\% & \textbf{22.7\%} & 21.7\% & 21.7\% & 1.4\,pp \\
\bottomrule
\end{tabular}
\caption{Best EM per root width $k_1$ at fixed budget $B{=}20$. For each $(\text{domain},k_1)$ cell we take the maximum EM among all swept TBS configurations with that root width that fit the budget. Bold marks the best $k_1$ per domain. 2D Maze and Maze3D are essentially $k_1$-insensitive (range $\le 1$\,pp); Sokoban requires wider roots ($+6.2$\,pp from $k_1{=}2$ to $k_1{=}8$). Strict-OOD Keys-and-Doors prefers a balanced $k_1{=}4$ versus BoN-20 at $0.7\%$.}
\label{tab:root_sensitivity}
\end{table}

\paragraph{Reading.}
The root-width-sensitivity range across domains exposes the operating mode of TBS in each:
\begin{itemize}
\item \textbf{Maze and Maze3D} are $k_1$-insensitive (range $\le 1$\,pp): TBS converges to nearly the same EM regardless of root width, and depth does the work.
\item \textbf{Sokoban} requires wider roots: a single-root TBS run ($k_1{=}2$) yields only $76.5\%$, while $k_1{=}8$ reaches $82.7\%$. Sokoban prefixes can be visually plausible but already strategically dead, so additional independent roots are required to escape shallow-rollback traps.
\item \textbf{Sliding Puzzle and Keys-Doors ID} prefer slightly wider roots ($k_1 \in \{4, 5, 8\}$ with $\le 2$\,pp spread).
\item \textbf{Strict-OOD Keys-and-Doors} prefers a balanced $k_1{=}4$, but every $k_1$ value yields $\ge 18.9\%$ EM versus BoN-20's $0.7\%$. The mechanism (prefix reuse), not the specific shape, is what matters.
\end{itemize}

\subsection{Depth--Width Search Shape}
\label{app:depth_width}

The matched-budget table above isolates the effect of $k_1$ at the single budget $B{=}20$. Here we show the full root-width versus budget interaction. For each domain and each root width $k_1 \in \{2, 4, 8\}$, we take the best EM achievable across all swept $(s, m)$ configurations satisfying $k_1 + s(m-1) = B$ at every available $B$. Lines only span the budgets where that $k_1$ was actually swept.

\begin{figure}[h]
\centering
\includegraphics[width=\textwidth]{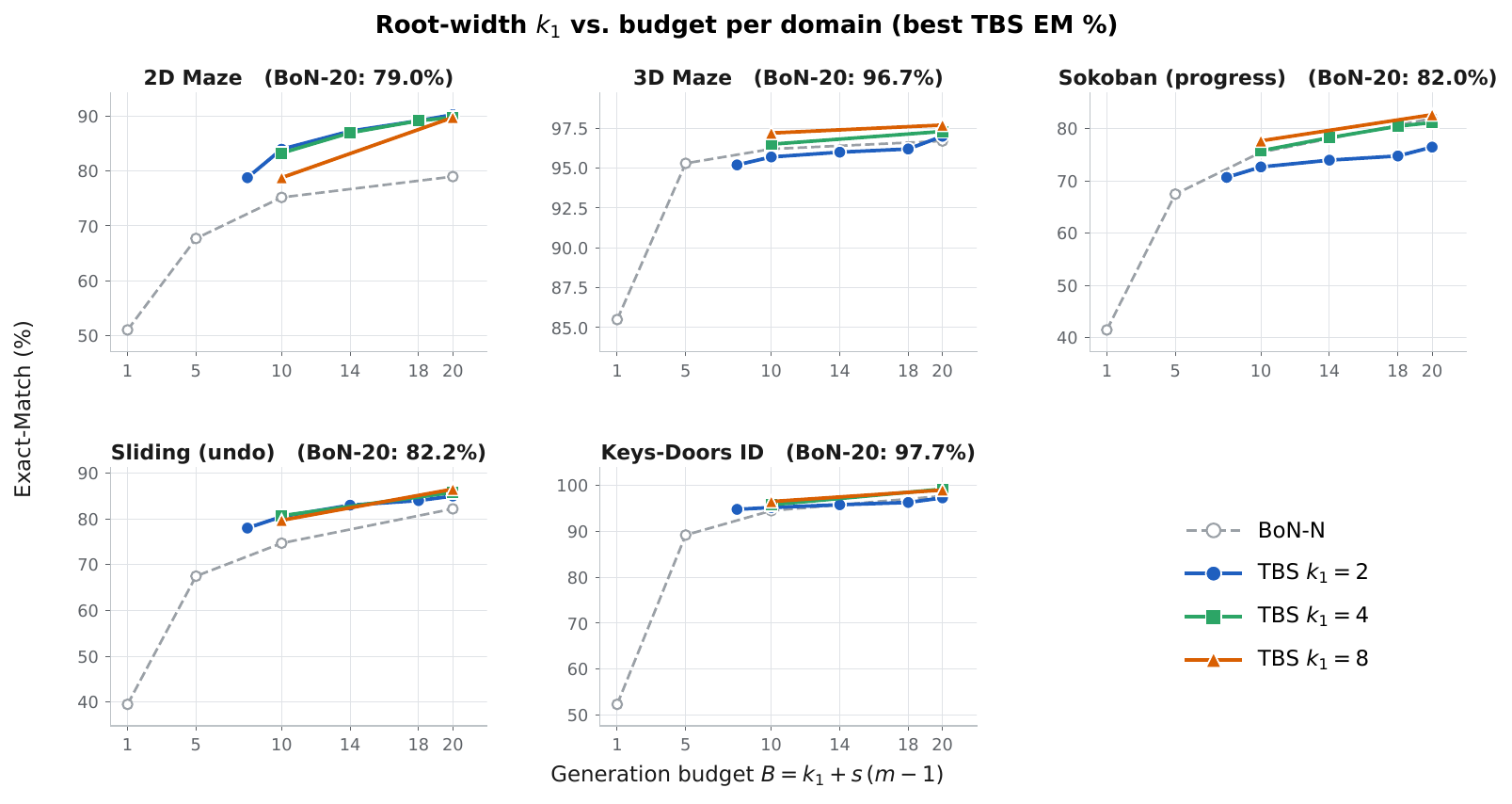}
\caption{Per-domain root-width $k_1$ versus generation budget $B$. Each panel shows the best TBS EM (\%) at every swept $(k_1, B)$ combination for that domain; dashed gray is the BoN-$N$ reference for $N \in \{1, 5, 10, 20\}$. \textbf{Sokoban} shows the strongest $k_1$ separation (wider roots dominate at every budget). \textbf{2D Maze} prefers $k_1{=}2$ (depth-friendly), with the narrow-and-deep curve sitting above wider variants at every budget. \textbf{3D Maze} and \textbf{Keys-Doors ID} are nearly $k_1$-insensitive at the saturation ceiling. Strict-OOD Keys-and-Doors is omitted from this plot because its absolute EM range (0--23\%) cannot be meaningfully compared to in-distribution scales; per-budget numbers are reported in Table~\ref{tab:root_sensitivity} and the strict-OOD path-cell breakdown in Table~\ref{tab:keys_ood_pathlen}.}
\label{fig:app_root_budget_curves}
\end{figure}

\paragraph{Cross-domain takeaway.}
The figure shows three regimes consistent with the per-domain analysis:
\begin{itemize}
\item \textbf{Depth-friendly (2D Maze):} $k_1{=}2$ dominates wider variants at every budget. Narrow-deep search converts each extra budget unit into prefix-reuse depth, which keeps producing new solves (cumulative EM by depth in Table~\ref{tab:cumulative_depth_maze} reaches $90.3\%$ at depth 6).
\item \textbf{Width-friendly (Sokoban):} $k_1{=}8$ dominates at every budget. Sokoban prefixes are strategically fragile, so wider roots are required to escape shallow-rollback traps before deeper search becomes useful.
\item \textbf{Saturated (3D Maze, Keys-Doors ID):} all three TBS curves cluster within $\le 2$\,pp of one another and well above the BoN reference. The mechanism helps but the operating ceiling limits visible separation across $k_1$.
\end{itemize}


\subsection{Verifier-Quality Ablations: Sokoban BFS-Aware and Sliding Undo-Required}
\label{app:verifier_quality_ablation}

The Sokoban and Sliding Puzzle headline TBS rows in the main paper use domain-specific verifier upgrades. Sokoban uses \texttt{SOKOBAN\_VERIFIER\_MODE=smart} with \texttt{SCORE\_MODE=progress}, which exposes a BFS dead-state and remaining-moves signal; without it, the verifier returns priority based purely on prefix length (the \emph{prefix-only} verifier). Sliding Puzzle uses \texttt{SLIDING\_SEMANTIC\_POLICY=undo\_required}, which marks any move whose only optimal next move is to undo the previous move as a \emph{semantic} failure and triggers rollback to the prior settled state; the alternative \texttt{SLIDING\_SEMANTIC\_POLICY=none} waits for a syntactic transition error. This appendix isolates how much each verifier upgrade contributes to TBS, holding generation, dataset, and search structure fixed.

\begin{figure}[h]
\centering
\includegraphics[width=\textwidth]{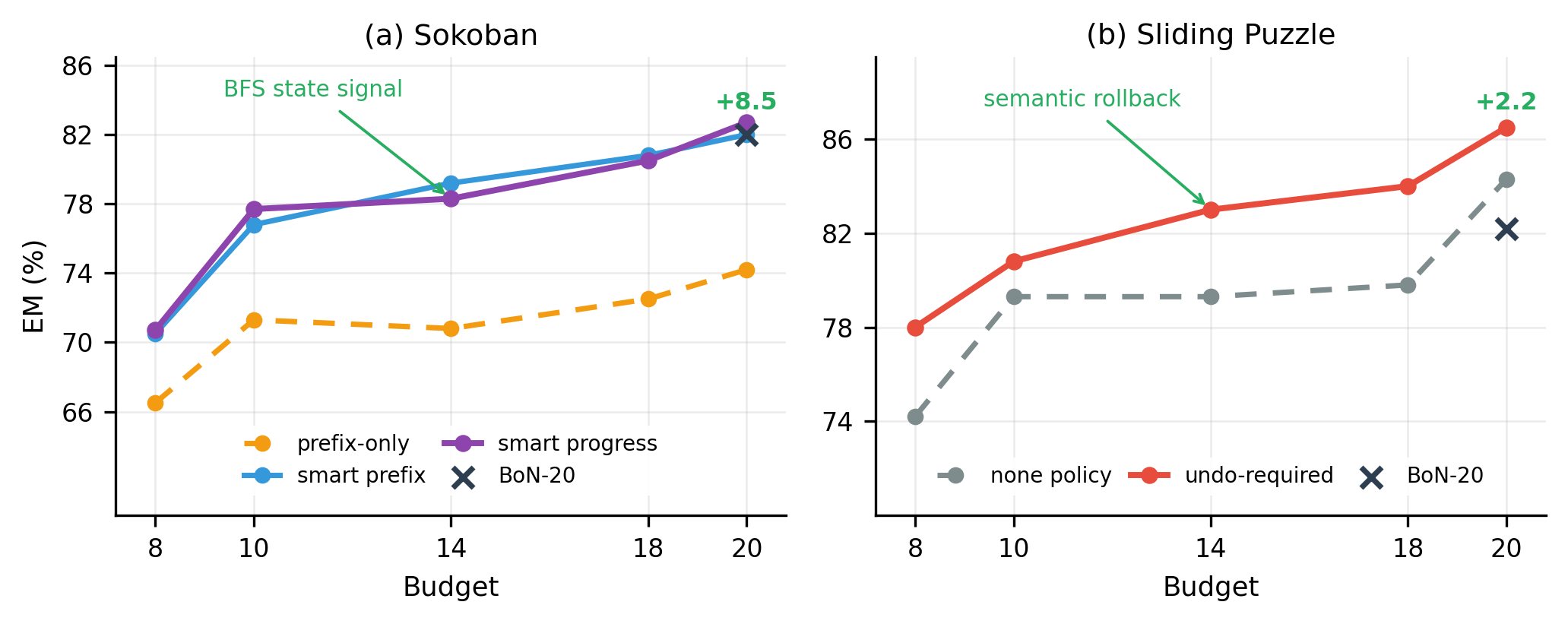}
\caption{\textbf{Verifier-quality ablations for state-aware restart.} \textbf{(a)} On Sokoban, the prefix-only verifier underperforms because long visual prefixes can already be strategically dead; adding BFS-based state verification and progress priority recovers the gap. \textbf{(b)} On Sliding Puzzle, undo-required semantic rollback improves TBS across budgets while the corresponding BoN-20 reference does not improve, showing that the gain comes from cleaner restart anchors rather than a more permissive terminal evaluator.}
\label{fig:verifier_ablation_panel}
\end{figure}

\paragraph{Sokoban: state-aware verification is necessary.}
Sokoban is the clearest verifier-quality ablation in the structured suite because visual prefix length is not the same as useful progress. A rollout can be visually legal for many frames while pushing the box into a dead state from which the target is unreachable. Conditioning on the longest prefix then preserves the wrong strategic state, and TBS repeatedly re-attempts from a prefix that should have been rejected.

We compare three variants on the same Sokoban. The \emph{prefix-only} variant ranks nodes by prefix length and does not use BFS dead-state information. The \emph{BFS prefix-search} variant uses the BFS-aware Sokoban verifier to reject unsolvable box states, but still uses the standard prefix-based priority. The \emph{progress-aware} variant is the headline setting: it ranks EM first, then BFS non-dead state, fewer BFS remaining moves, BFS progress, and finally the prefix fallback.

\begin{table}[h]
\centering\small
\setlength{\tabcolsep}{4pt}
\begin{tabular}{lcccc}
\toprule
TBS config & Budget & Prefix-only verifier & Progress-aware verifier & $\Delta$ verifier \\
\midrule
$k_1{=}2,s{=}3,m{=}3$        &  8 & $399/600 = 66.5\%$ & $424/600 = 70.7\%$ & $+4.2$\,pp \\
$k_1{=}2,s{=}2,m{=}5$        & 10 & $401/600 = 66.8\%$ & $436/600 = 72.7\%$ & $+5.9$\,pp \\
$k_1{=}2,s{=}2,m{=}7$        & 14 & $409/600 = 68.2\%$ & $444/600 = 74.0\%$ & $+5.8$\,pp \\
$k_1{=}2,s{=}2,m{=}9$        & 18 & $414/600 = 69.0\%$ & $449/600 = 74.8\%$ & $+5.8$\,pp \\
$k_1{=}4,s{=}2,m{=}9$        & 20 & $438/600 = 73.0\%$ & $487/600 = 81.2\%$ & $+8.2$\,pp \\
wider $k_1{=}2,s{=}2,m{=}10$ & 20 & $415/600 = 69.2\%$ & $453/600 = 75.5\%$ & $+6.3$\,pp \\
wider $k_1{=}2,s{=}3,m{=}7$  & 20 & $418/600 = 69.7\%$ & $459/600 = 76.5\%$ & $+6.8$\,pp \\
wider $k_1{=}5,s{=}3,m{=}6$  & 20 & $431/600 = 71.8\%$ & $492/600 = 82.0\%$ & $+10.2$\,pp \\
wider $k_1{=}8,s{=}2,m{=}7$  & 20 & $445/600 = 74.2\%$ & $496/600 = 82.7\%$ & $+8.5$\,pp \\
\bottomrule
\end{tabular}
\caption{Sokoban verifier-quality ablation under matched TBS configurations: prefix-only verification versus progress-aware verification (BFS dead-state, remaining-moves, and progress). The progress-aware verifier contributes $+4.2$ to $+10.2$\,pp across all 9 matched configurations spanning budgets 8 to 20, with the gap widening at higher budgets and wider roots. Dataset, generation model, prompt, and budget are held fixed; only \texttt{SOKOBAN\_VERIFIER\_MODE} and \texttt{SCORE\_MODE} change.}
\label{tab:sokoban_smart_vs_dumb}
\end{table}

\paragraph{Reading.}
The prefix-only variant is far below BoN-20, so naive temporal branching is actively harmful on Sokoban. The BFS prefix-search variant recovers to BoN-20 parity, and progress-aware ranking adds the small headline gain. This matches the per-domain restart-quality analysis: Sokoban's deeper TBS expansions are dominated by tracking and dead-state failures, so a state-aware priority function is required to escape shallow-rollback traps and rank non-dead BFS states above visually plausible but unsolvable prefixes.

The main lesson is not that Sokoban gives a large TBS gain---it does not. Rather, prefix reuse is only as good as the process signal defining the prefix. Maze prefixes are usually reusable because a valid route prefix remains useful. Sokoban prefixes can be \emph{poisoned}: the box may be in a legal but strategically dead configuration. State-aware verification is therefore required before prefix conditioning becomes a helpful search primitive.

\paragraph{Sliding Puzzle: undo-required semantic rollback.}
Sliding Puzzle has a different verifier failure mode. A generated tile move can be locally legal under the blank-cell rule while still moving away from the shortest path to the goal board. Under \texttt{SLIDING\_SEMANTIC\_POLICY=none}, the verifier waits until a syntactic error appears (illegal transition, duplicate tile, wrong settled board), by which point the restart prefix can already be semantically poisoned. Under \texttt{SLIDING\_SEMANTIC\_POLICY=undo\_required}, the verifier checks each settled board state using the puzzle rules and goal board: if the only optimal next move is to undo the previous move, the previous move is treated as the semantic failure and TBS rolls back to the prior settled state. This gives TBS a cleaner restart anchor without providing the correct future continuation.

\begin{table}[h]
\centering\small
\setlength{\tabcolsep}{4pt}
\begin{tabular}{lcccc}
\toprule
TBS config & Budget & None policy & Undo-required & $\Delta$ verifier \\
\midrule
$k_1{=}2,s{=}3,m{=}3$        &  8 & $445/600 = 74.2\%$ & $468/600 = 78.0\%$ & $+3.8$\,pp \\
$k_1{=}2,s{=}2,m{=}5$        & 10 & $467/600 = 77.8\%$ & $483/600 = 80.5\%$ & $+2.7$\,pp \\
$k_1{=}2,s{=}2,m{=}7$        & 14 & $476/600 = 79.3\%$ & $498/600 = 83.0\%$ & $+3.7$\,pp \\
$k_1{=}2,s{=}2,m{=}9$        & 18 & $479/600 = 79.8\%$ & $504/600 = 84.0\%$ & $+4.2$\,pp \\
$k_1{=}4,s{=}2,m{=}9$        & 20 & $495/600 = 82.5\%$ & $515/600 = 85.8\%$ & $+3.3$\,pp \\
wider $k_1{=}2,s{=}2,m{=}10$ & 20 & $480/600 = 80.0\%$ & $506/600 = 84.3\%$ & $+4.3$\,pp \\
wider $k_1{=}2,s{=}3,m{=}7$  & 20 & $485/600 = 80.8\%$ & $510/600 = 85.0\%$ & $+4.2$\,pp \\
wider $k_1{=}5,s{=}3,m{=}6$  & 20 & $498/600 = 83.0\%$ & $519/600 = 86.5\%$ & $+3.5$\,pp \\
wider $k_1{=}8,s{=}2,m{=}7$  & 20 & $506/600 = 84.3\%$ & $519/600 = 86.5\%$ & $+2.2$\,pp \\
\bottomrule
\end{tabular}
\caption{Sliding Puzzle verifier-quality ablation: \texttt{none} versus \texttt{undo\_required} semantic policy across all budgets and search configurations. The undo-required policy slightly hurts BoN at every budget ($-0.8$ to $-1.6$\,pp) because it changes how the verifier classifies borderline rollouts. It consistently helps TBS by $+2.2$ to $+4.3$\,pp across all 9 matched configurations spanning budgets 8 to 20. The asymmetry isolates a TBS-specific contribution: cleaner restart anchors only matter once you are restarting from them. The gain trends slightly upward with search depth (e.g., $+2.7$ at $m{=}5$, $+4.2$ at $m{=}9$ at fixed $k_1{=}2$), consistent with deeper search benefiting more from cleaner anchors.}
\label{tab:sliding_undo_vs_none}
\end{table}

\paragraph{Reading.}
The undo-required policy consistently helps TBS by $+2.2$ to $+4.3$\,pp. This is the cleanest evidence that the verifier upgrade interacts with restart specifically: the smarter verifier produces better restart anchors, which has no effect on a method that does not use restart anchors (BoN) but a measurable effect on a method that does (TBS). It also explains why the headline Sliding TBS-20 row in the main paper uses $k_1{=}5, s{=}3, m{=}6$ with undo-required: the wider-root shape works in concert with the cleaner-anchor verifier.

\paragraph{Cross-domain takeaway.}
Both ablations are consistent with the same conclusion. Where the basic prefix-length priority already aligns with structurally sound restart anchors (Maze), TBS gains immediately from prefix conditioning. Where basic prefix-length priority returns visually or semantically corrupted anchors (Sokoban, Sliding), a domain-specific verifier upgrade is the operational path to TBS-style gains: BFS-progress for Sokoban, undo-required semantics for Sliding.

\subsection{Search-Policy Ablation: Greedy, Beam-2, and Random Frontiers}
\label{app:search_policy_ablation}
The main paper compares TBS against matched-budget root-only BoN. To isolate whether the gain comes merely from any prefix reuse or from the priority-frontier search policy, we run three additional prefix-reuse controls on the hard structured splits. All controls use the same generator, prompt, verifier, terminal evaluator, and per-domain TBS search shape $(k_1,s,m)$ as the corresponding TBS row. Only the frontier policy changes:

\begin{itemize}
\item \textbf{Greedy:} priority pop with frontier cap $1$. This is single-lineage prefix repair: after each expansion, only the best verified anchor is retained.
\item \textbf{Beam-2:} priority pop with frontier cap $2$. This retains the best two verified anchors after each expansion, testing whether one alternate prefix is enough.
\item \textbf{Random:} uncapped frontier with random pop. This keeps frontier diversity but disables the verifier priority at the pop step.
\end{itemize}

\begin{table}[h]
\centering\small
\setlength{\tabcolsep}{4pt}
\begin{tabular}{lccccc}
\toprule
Domain & BoN-20 & Greedy & Beam-2 & Random & TBS-20 \\
\midrule
2D Maze Hard & 60.0 & 61.0 & 78.3 & 74.7 & \textbf{81.0} \\
Maze3D Hard & 93.7 & 95.0 & \textbf{95.3} & 95.0 & \textbf{95.3} \\
Sokoban Hard, progress-aware & 78.3 & 74.7 & 77.3 & 78.7 & \textbf{79.7} \\
Sliding Hard, undo-required & 81.3 & 83.7 & 87.3 & 84.0 & \textbf{88.7} \\
Keys-Doors ID Hard & 97.7 & 96.0 & 97.3 & 98.7 & \textbf{99.0} \\
Keys-Doors strict-OOD & 0.7 & 6.7 & 19.7 & 12.7 & \textbf{22.7} \\
\bottomrule
\end{tabular}
\caption{Search-policy ablation at matched headline budget $B{=}20$ on hard / strict-OOD structured splits. Greedy, Beam-2, Random, and TBS use the same per-domain TBS-20 shape; only the frontier memory and pop rule change. All entries are exact-match percentages over $n{=}300$ samples. TBS is best or tied on every split.}
\label{tab:policy_ablation_b20}
\end{table}

\paragraph{Reading at $B{=}20$.}
Greedy prefix repair is not enough: it is far below TBS on 2D Maze ($61.0$ vs.\ $81.0$), Sliding ($83.7$ vs.\ $88.7$), Sokoban ($74.7$ vs.\ $79.7$), Keys-ID ($96.0$ vs.\ $99.0$), and strict-OOD Keys ($6.7$ vs.\ $22.7$). Beam-2 recovers much of the gap because it preserves one alternate restart anchor, but full TBS remains best or tied on all splits: $+2.7$\,pp over Beam-2 on Maze, $+2.4$\,pp on Sokoban, $+1.4$\,pp on Sliding, $+1.7$\,pp on Keys-ID, and $+3.0$\,pp on strict-OOD Keys. Random sometimes beats Greedy by avoiding single-anchor over-commitment, but it remains below TBS on every split with headroom, showing that the verifier priority is useful in addition to frontier diversity.

\paragraph{Conclusion.}
The policy ablation refines the claim. The dominant mechanism is verified temporal prefix reuse, but the full priority frontier is not redundant. Greedy single-lineage repair over-commits. Beam-2 is a strong compact approximation, but full TBS is best or tied at the headline budget across all six structured splits.

\section{Target-Bench Navigation Details}
\label{app:targetbench}

\subsection{Official benchmark score and metric definitions}

Target-Bench reports five trajectory metrics combined into a single overall score:
\begin{equation}
S = 0.05\,e^{-\mathrm{ADE}}
+ 0.10\,e^{-\mathrm{FDE}}
+ 0.10\,(1{-}\mathrm{MR}/100)
+ 0.65\,\mathrm{SE}\cdot\mathrm{AC},
\label{eq:appendix_score}
\end{equation}
where ADE / FDE are average and final L2 displacement errors against the GT trajectory (interpolated to GT length), MR is the percentage of interpolated points farther than 1\,m from GT, SE is a soft endpoint score $\exp(-\text{endpoint\_error}^2 / 2\sigma^2)$ with $\sigma{=}0.5$, and AC is an approach-consistency corridor score using GT reference points with $\sigma_{\min}{=}0.5, \sigma_{\max}{=}1.5, \beta{=}0.25, \gamma{=}5.0$.

Trajectories are decoded from generated videos via VGGT, anchored to the first frame pose, and converted to benchmark metric space using the calibration $\lambda = d_{\text{real}} / d_{\text{pred}}$. The reported numbers are the official benchmark overall score of the selected generated video.

\subsection{Predictor details}
The Target-Bench verifier is a learned trajectory predictor that maps the current frame and prompt to a metric-space future path. The predictor uses V-JEPA2 ViT-G/384 image tokens and text embeddings, followed by a fusion image-text transformer with 4 layers, 8 heads, hidden dimension 512, and dropout 0.1. It predicts 9 future trajectory points in benchmark metric space. Training uses 100 scenarios with a weighted point loss $[1.0,\ 1.0,\ 1.0,\ 1.2,\ 1.5,\ 1.8,\ 2.2,\ 2.6,\ 3.0]$ and a late-$x$ weight of 0.2 starting at point index 5.

At test time, a generated video is decoded with VGGT into a trajectory and compared against the predicted path at the 9 prefix points. We use a bell-shaped per-axis tolerance schedule with $\tau_{\text{base}}{=}0.15$\,m at the prefix endpoints rising to $0.50$\,m near the middle, scaled by $\tau_x$/$\tau_y$. A prefix point is invalid when $|dx|>\tau_x$ or $|dy|>\tau_y$. The first invalid prefix point defines the restart anchor, with restart backoff 1. TBS-3 uses $k_1{=}2,s{=}1,m{=}2$ with  TBS-5 uses $k_1{=}4,s{=}1,m{=}2$.

\section{RoboTwin Manipulation Details}
\label{app:robotwin}

\subsection{Pipeline}

A generated video is decoded into a 14-D RoboTwin/Aloha action sequence by an inverse dynamics model and replayed in the RoboTwin simulator. Verifier inputs are the simulator's per-step state: gripper open/close events, object displacement, object--target distance, task-specific progress, and the official task predicate \texttt{check\_success()}.

\begin{itemize}
\item Generator: released VIDAR / Wan-style TI2V~\citep{feng2025vidar}.
\item Action decoder: released VIDAR IDM~\citep{feng2025vidar}.
\item Action format: 14-D RoboTwin/Aloha sequences.
\item Observation: stitched 3-view RGB image (head + left/right wrist cameras).
\item Semantic FPS: 8.
\item Segment length: 61 frames per generation (1 conditioning + 60 generated), with optional 61+61 continuation when the first segment does not yet contain a goal-attempt window.
\item Search budget is counted at the rollout/branch level.
\item TBS-3 uses $k_1{=}1,s{=}1,m{=}3$; TBS-5 uses $k_1{=}1,s{=}1,m{=}5$.
\end{itemize}

\subsection{First-goal-attempt window}

The verifier identifies a \emph{first goal attempt}: a gripper close--open window co-occurring with object motion toward the task target and a release/settle event near the goal pose. Detection uses simulator state, IDM-decoded actions, gripper timing, and object--target distance. When a root rollout fails after a meaningful first attempt, TBS restarts from the preceding gripper-opening boundary---not from the initial observation---so the pre-grasp setup motion is preserved and compute is concentrated on repairing the failed manipulation.

This avoids two failure modes of naive restart: (i) crediting repeated retries after the first failed attempt as separate solves, and (ii) discarding successful pre-grasp motion that took several seconds to produce.

\subsection{The 10-task progress-aware subset}

\begin{table}[h]
\centering\small
\begin{tabular}{lll}
\toprule
Task & Type & Restart cue \\
\midrule
\texttt{place\_bread\_skillet} & place & release near skillet \\
\texttt{stack\_blocks\_two} & stack & failed stack/release \\
\texttt{place\_burger\_fries} & place & release near target \\
\texttt{place\_shoe} & place & object-target release \\
\texttt{stack\_bowls\_two} & stack & failed bowl stack/release \\
\texttt{place\_a2b\_right} & move/place & gripper-open near target \\
\texttt{place\_a2b\_left} & move/place & gripper-open near target \\
\texttt{place\_fan} & place & release near fan target \\
\texttt{place\_mouse\_pad} & place & release near pad target \\
\texttt{place\_object\_scale} & place & release near scale \\
\bottomrule
\end{tabular}
\caption{RoboTwin 10-task progress-aware subset used in the paper.}
\label{tab:robotwin_tasks}
\end{table}

\paragraph{Selection criteria.}
The subset is chosen to test the mechanism TBS is designed for: branching from verified partial progress. Tasks require progress-aware manipulation rather than one-shot contact triggering, a clear first-goal-attempt window detectable from simulator state, visible sub-goal structure (reach/grasp/move/place/stack/release).

\paragraph{Evaluation protocol.}
We run $10$ evaluation episodes per task across the $10$ tasks listed above. Each cell is evaluated under two scene-randomization regimes: \emph{clean}, which preserves the canonical RoboTwin initial state, and \emph{randomized}, an out-of-distribution split that perturbs the visual background and object placement relative to the conditions seen by the generator. Both regimes use the same task list, the same per-cell episode count, and the same matched-budget BoN versus TBS comparison.

\begin{table}[h]
\centering\small
\begin{tabular}{llccc}
\toprule
Setting & Method & Budget & Successes & Success rate \\
\midrule
Clean & BoN & 3 & $46/100$ & $46\%$ \\
Clean & TBS & 3 & $50/100$ & $50\%$ \\
Clean & BoN & 5 & $49/100$ & $49\%$ \\
Clean & TBS & 5 & $53/100$ & $53\%$ \\
\midrule
Randomized & BoN & 3 & $15/100$ & $15\%$ \\
Randomized & TBS & 3 & $16/100$ & $16\%$ \\
Randomized & BoN & 5 & $17/100$ & $17\%$ \\
Randomized & TBS & 5 & $18/100$ & $18\%$ \\
\bottomrule
\end{tabular}
\caption{RoboTwin selected-subset results. In clean scenes, TBS gains $+4$ successful episodes over matched-budget BoN at both budgets. In randomized scenes, absolute performance is lower and the gain is smaller because fewer root rollouts produce repairable verified prefixes.}
\label{tab:robotwin_results}
\end{table}

\section{Limitations and Failure Cases}
\label{app:limitations}

\paragraph{TBS requires a useful verified prefix.}
TBS amplifies a base model's local competence: it can stitch verified partial trajectories into longer correct ones. It cannot manufacture competence the base model does not have. When the base rollout fails before producing meaningful progress (RoboTwin randomized setting, Sliding hard with semantic prefix corruption, Maze3D 24-grid where the template tracker loses the agent on long horizons), TBS has little to reuse and the gain shrinks toward zero.

\paragraph{Verifier mislocalization.}
TBS depends on the verifier returning an approximately correct restart anchor. The Sokoban smart verifier exposes the limit case: deeper TBS expansions in Sokoban are dominated by tracking failures, meaning the verifier rolls back only a few frames before re-failing on visually unstable prefixes. The width-versus-depth ablation (Section~\ref{app:depth_width}) shows that wider roots, not deeper search, mitigate this for Sokoban. Sliding undo-required has a similar structure: when the verifier mis-detects the wrong-board state, TBS inherits the corrupted prefix and re-fails.

\paragraph{Learned process supervision is the bottleneck on Target-Bench.}
The Target-Bench gain is positive at both matched budgets, but the remaining headroom depends on the quality of the learned process signal. The synthetic-noise ablation (Section~\ref{app:targetbench}) shows that TBS does not collapse under moderate process-hint noise but does converge with BoN at the strongest perturbation tested ($q{=}0.3$

, $\sim 0.7$\,m mean ADE noise).

\paragraph{Domain saturation.}
On 3D Maze, in-distribution Keys-and-Doors, and most Sokoban splits, BoN-1 already solves a large fraction of problems. TBS matches or slightly exceeds BoN at matched budget on these but does not produce large gains. We treat them as null controls for downside risk rather than as evidence for TBS.

\paragraph{Compute overhead.}
At budget $B$, TBS requires the same nominal generation count as BoN-$B$, but the verifier must run on each rollout (root and child). For exact symbolic verifiers this is negligible; for the learned Target-Bench verifier and the simulator-replay RoboTwin verifier the overhead is measurable but small relative to the cost of video generation.

\vspace{-5.5cm}
\section{Visual Examples}
\label{app:video_examples}
\vspace{-3.5cm}

\begin{figure}[H]
    \centering
    \begin{subfigure}[t]{0.48\linewidth}
        \centering
        \includegraphics[width=\linewidth]{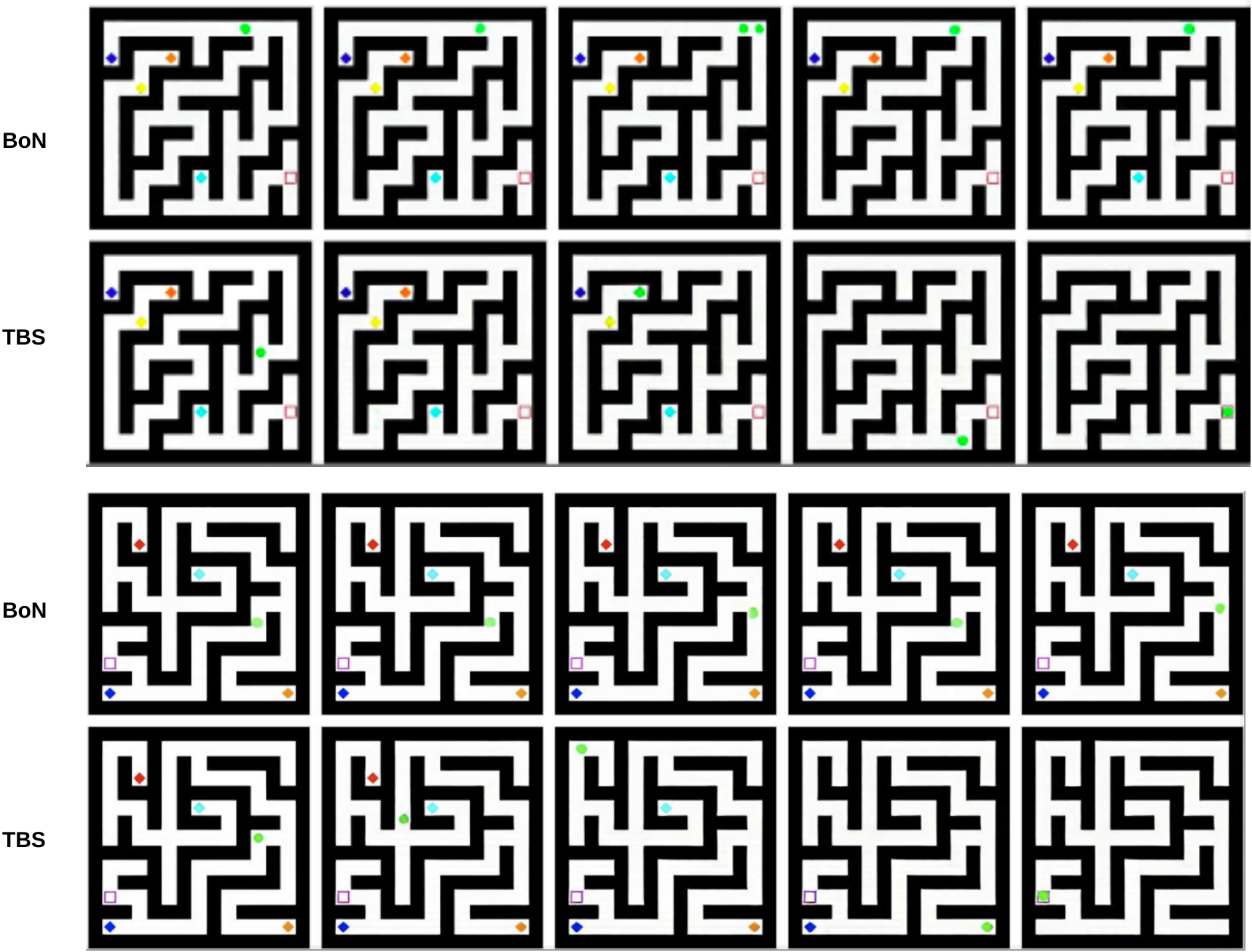}
        \caption{Example 1}
        \label{fig:exp1}
    \end{subfigure}
    \hfill
    \begin{subfigure}[t]{0.48\linewidth}
        \centering
        \includegraphics[width=\linewidth]{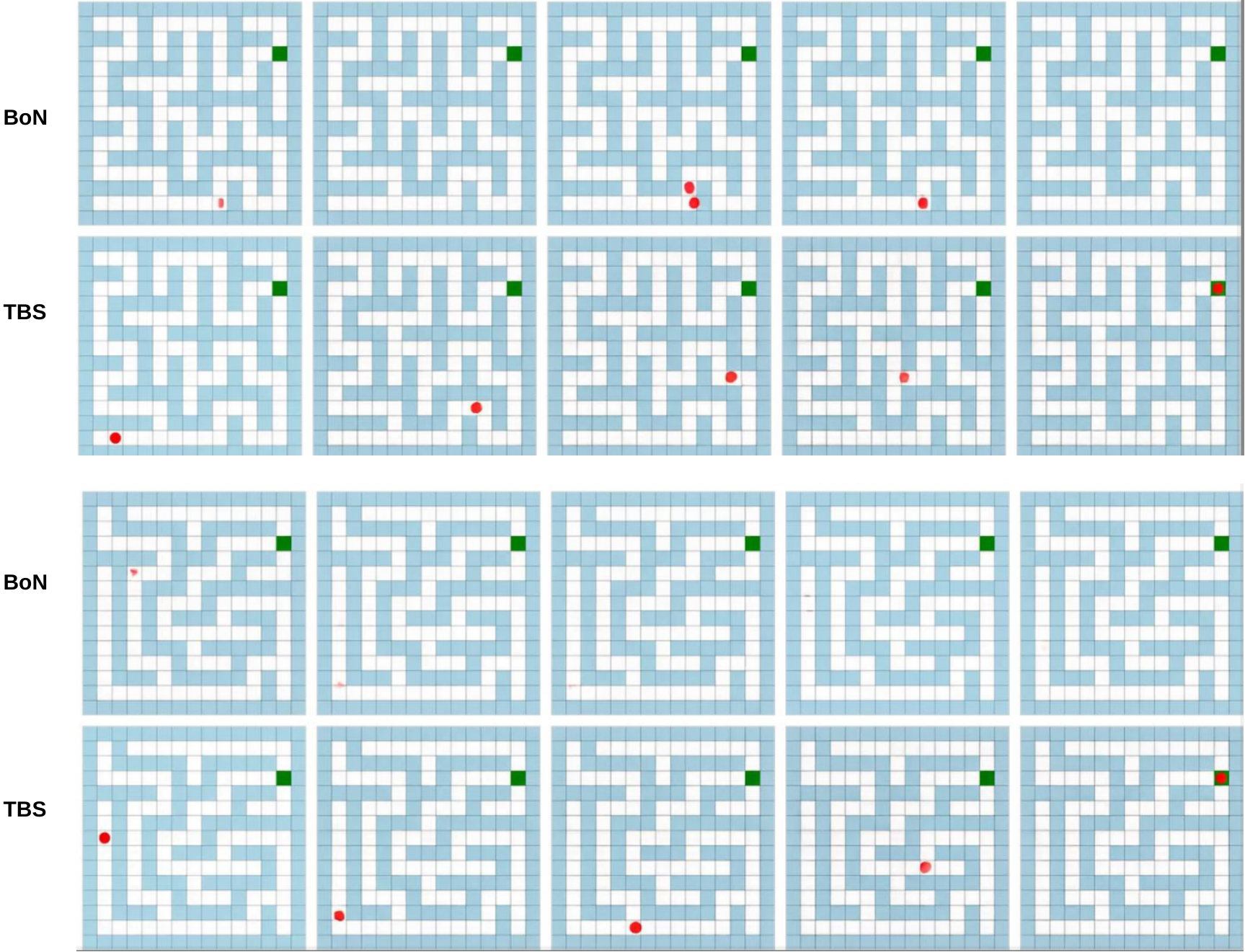}
        \caption{Example 2}
        \label{fig:exp2}
    \end{subfigure}

    \vspace{0.6em}

    \begin{subfigure}[t]{0.48\linewidth}
        \centering
        \includegraphics[width=\linewidth]{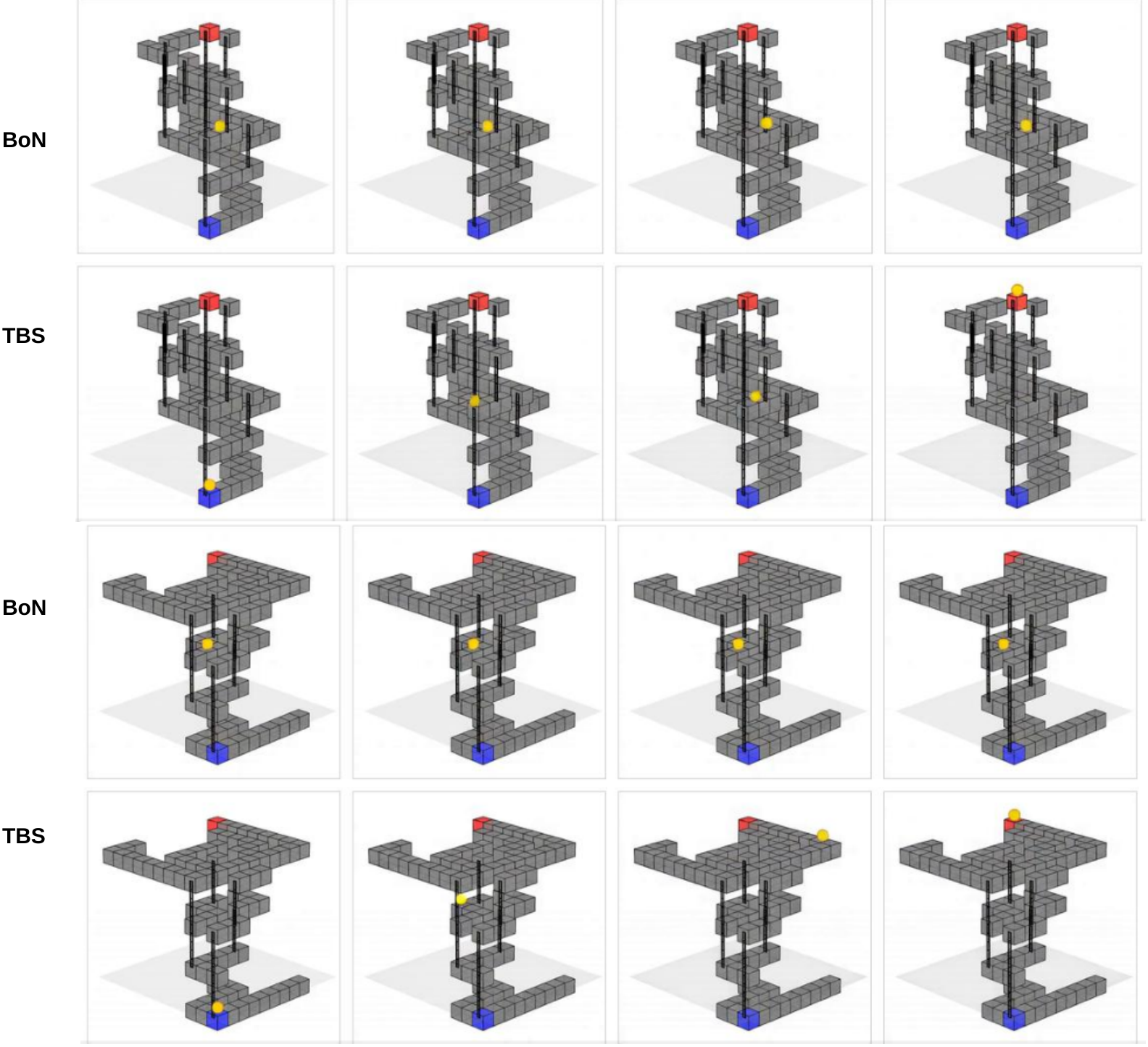}
        \caption{Example 3}
        \label{fig:exp3}
    \end{subfigure}
    \hfill
    \begin{subfigure}[t]{0.48\linewidth}
        \centering
        \includegraphics[width=\linewidth]{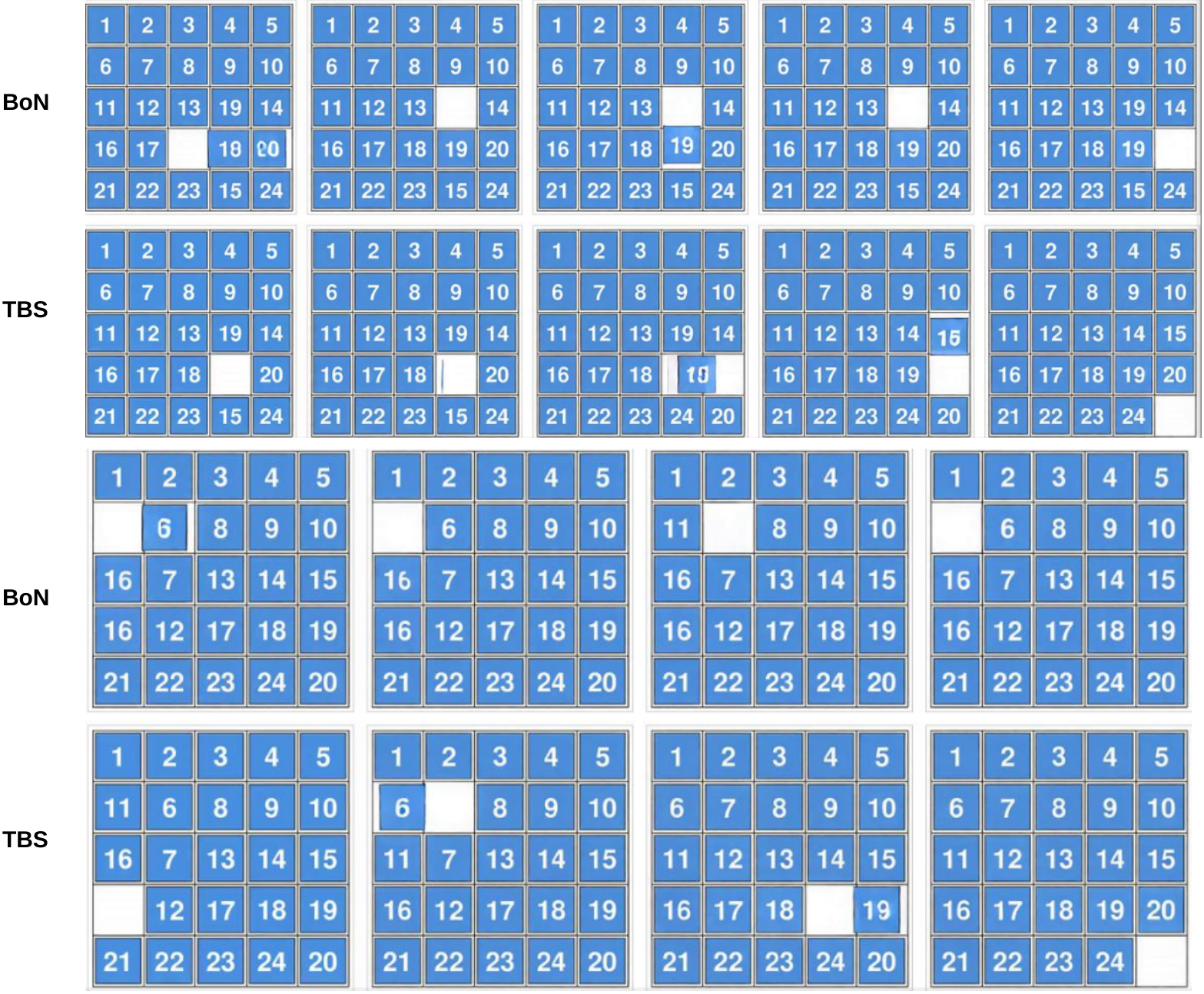}
        \caption{Example 4}
        \label{fig:exp4}
    \end{subfigure}
    \caption{Visual examples from the experiments, part 1.}
    \label{fig:visual_examples_part1}
\end{figure}

\begin{figure}[H]
    \centering
    \begin{subfigure}[t]{0.85\linewidth}
        \centering
        \includegraphics[width=\linewidth]{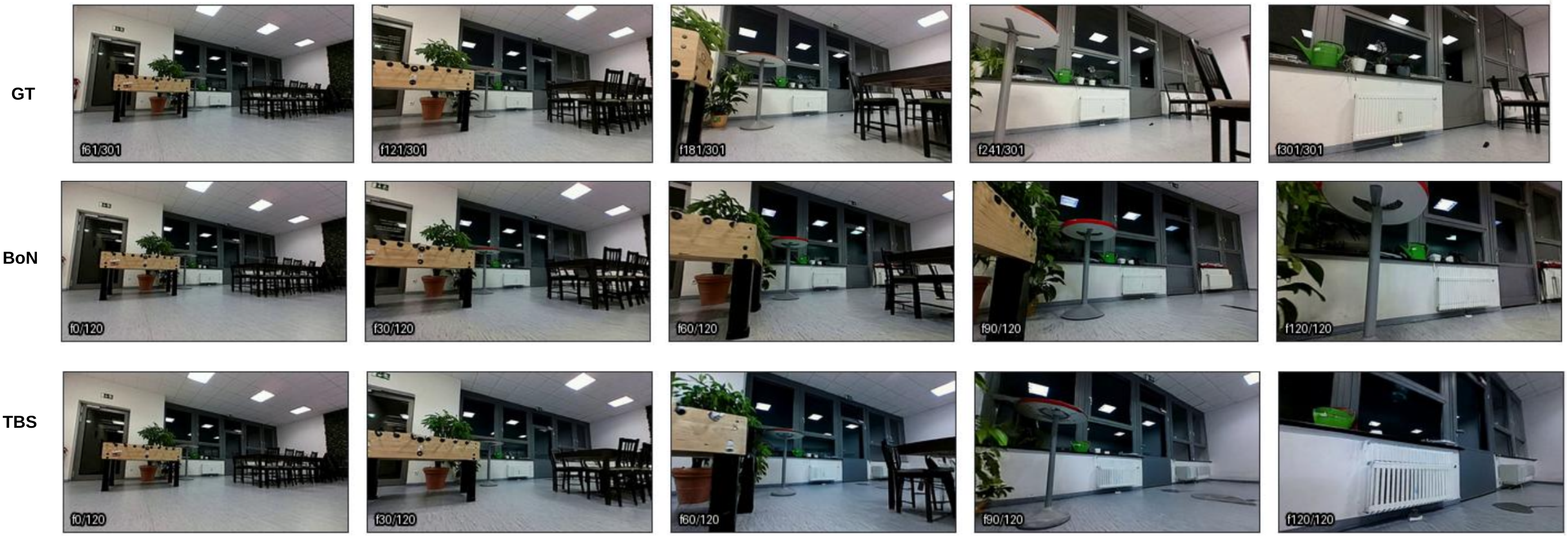}
        \caption{Example 5}
        \label{fig:exp5}
    \end{subfigure}

    \vspace{0.8em}

    \begin{subfigure}[t]{0.85\linewidth}
        \centering
        \includegraphics[width=\linewidth]{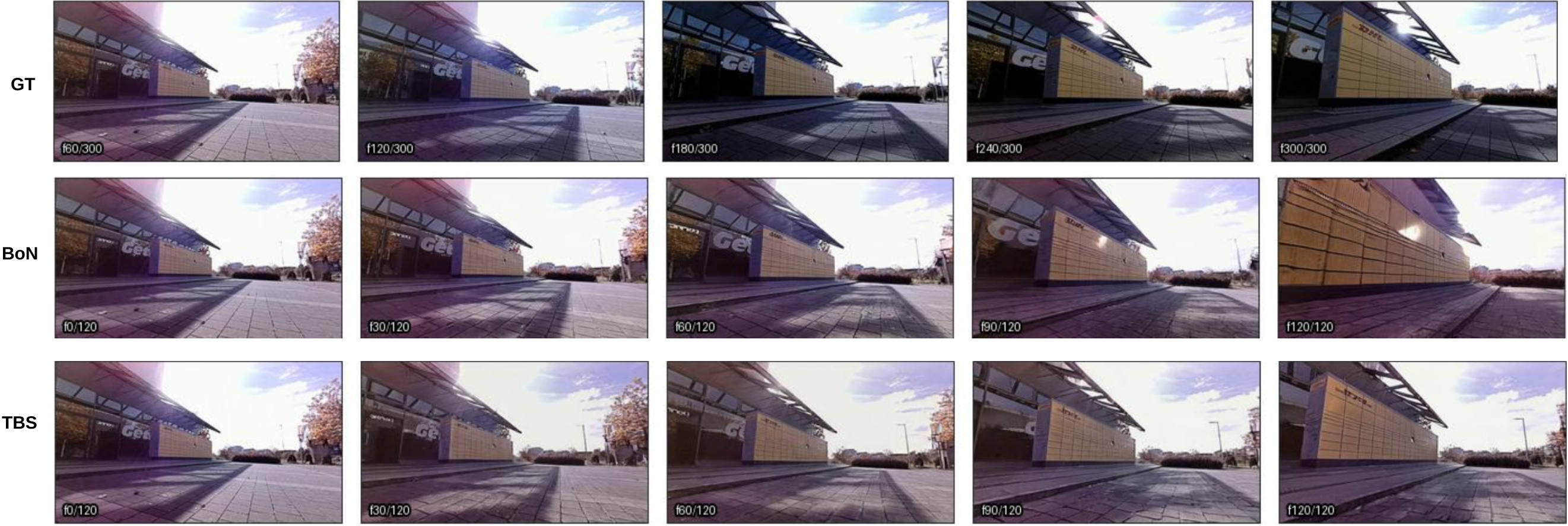}
        \caption{Example 6}
        \label{fig:exp6}
    \end{subfigure}

    \vspace{0.8em}

    \begin{subfigure}[t]{0.85\linewidth}
        \centering
        \includegraphics[width=\linewidth]{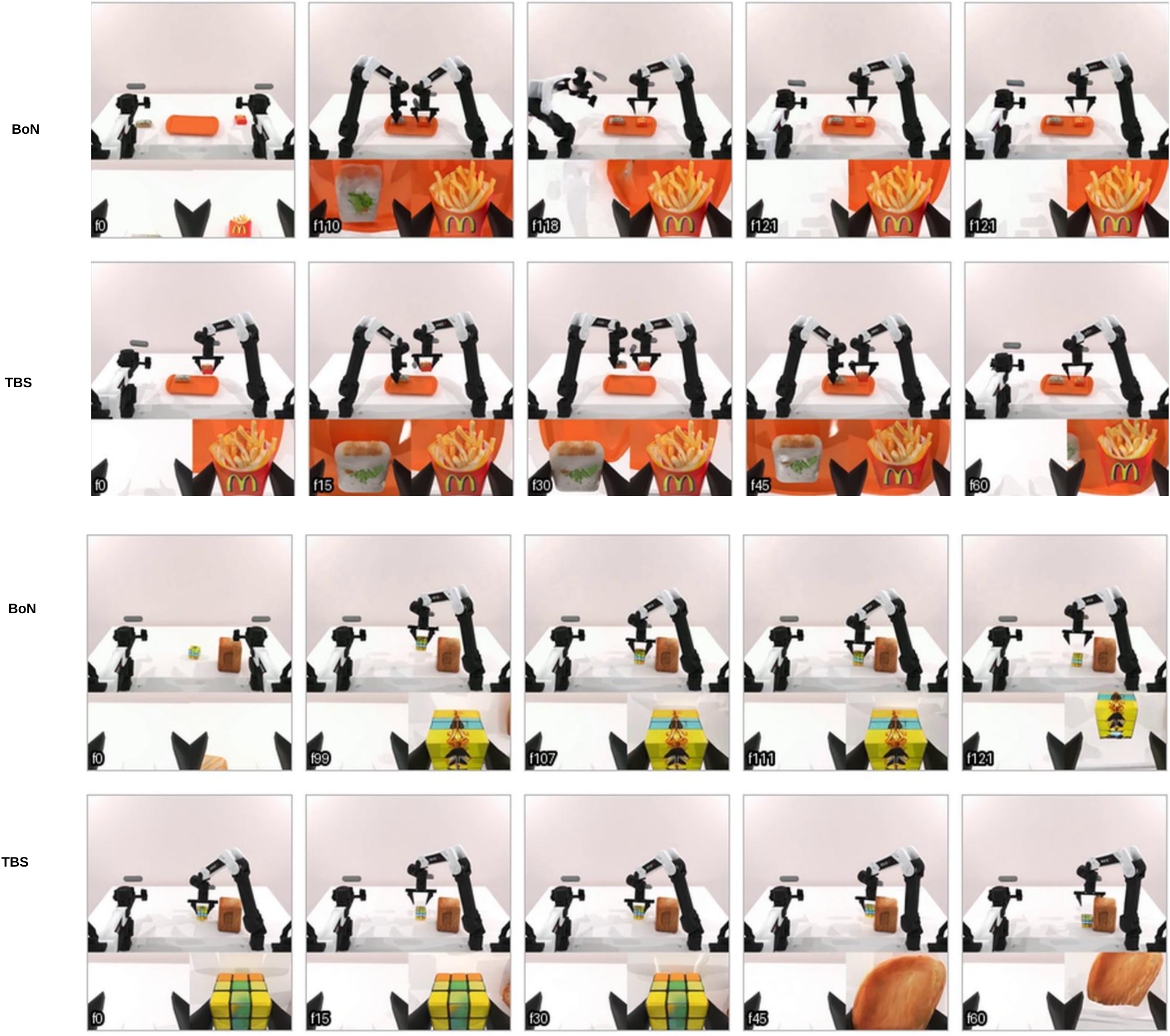}
        \caption{Example 7}
        \label{fig:exp7}
    \end{subfigure}
    \caption{Visual examples from the experiments, part 2.}
    \label{fig:visual_examples_part2}
\end{figure}

\clearpage

\newpage

\end{document}